\documentclass[10pt,twocolumn,letterpaper]{article}

\usepackage{3dv}
\usepackage{times}
\usepackage{epsfig}
\usepackage{graphicx}
\usepackage{amsmath}
\usepackage{amssymb}

\usepackage{arydshln}
\usepackage[dvipsnames]{xcolor}
\usepackage{soul}
\usepackage{url}
\usepackage{booktabs}
\usepackage{multirow}
\usepackage{float}
\usepackage{dblfloatfix}
\usepackage{wrapfig}
\usepackage{authblk}
\usepackage[pagebackref=true,breaklinks=true,letterpaper=true,colorlinks,bookmarks=false]{hyperref}

\threedvfinalcopy % *** Uncomment this line for the final submission

\def\threedvPaperID{133} % *** Enter the 3DV Paper ID here

\ifthreedvfinal\fi
\begin{document}

%Custom Commands
\newcommand{\miss}{\textcolor{red}{REF}}
\newcommand{\due}[1]{{\sethlcolor{red}\hl{#1}}}
\newcommand{\unsure}[1]{{\textcolor{Cerulean}{\textit{#1} (unsure)}}}

\newcommand{\HR}[1]{{\textcolor{magenta}{[\textbf{HR:} #1]}}}
\newcommand{\WX}[1]{{\textcolor{purple}{[\textbf{WX:} #1]}}}
\newcommand{\DC}[1]{{\textcolor{orange}{[\textbf{Dan:} #1]}}}
\newcommand{\CT}[1]{{\textcolor{red}{[\textbf{CT:} #1]}}}
\newcommand{\DM}[1]{{\textcolor{blue}{[\textbf{DM:} #1]}}}
\newcommand{\TODO}[1]{{\textcolor{red}{[TODO: #1]}}}
\newcommand{\change}[1]{{{#1}}}
\newcommand{\ignore}[1]{{}}

\def\mysection#1#2{\section{#1}\label{sec:#2}}
\def\mysubsection#1#2{\subsection{#1}\label{sec:#2}}
\def\mysubsubsection#1#2{\subsubsection{#1}\label{sec:#2}}

\def\figurePath{Figures/}
\newcommand{\myfigure}[3]{\begin{figure}\centering\includegraphics*[scale = #3]{\figurePath#1}\caption{#2}\label{fig:#1}\vspace{-0.5cm}\end{figure}}
\newcommand{\mywrapfigure}[3]{\begin{wrapfigure}{r}{0.35\textwidth}\centering\includegraphics*[scale = #3]{\figurePath#1}\caption{#2}\label{fig:#1}\end{wrapfigure}}
\newcommand{\mywrapfigurealt}[3]{\begin{wrapfigure}{r}{0.5\textwidth}\centering\includegraphics*[scale = #3]{\figurePath#1}\caption{#2}\label{fig:#1}\end{wrapfigure}}
\newcommand{\mysubfigure}[7]{\begin{figure}[h!]\centering\begin{subfigure}{.48\textwidth}\centering\includegraphics*[scale = #3]{\figurePath#1}\caption{#2}\label{fig:#1}\end{subfigure}\hfill\begin{subfigure}{.48\textwidth}\centering\includegraphics*[scale = #6]{\figurePath#4}\caption{#5}\label{fig:#4}\end{subfigure}\caption{#7}\end{figure}}

\newcommand{\refSec}[1]{Sec.~\ref{sec:#1}}
\newcommand{\refFig}[1]{Fig.~\ref{fig:#1}}
\newcommand{\refEq}[1]{Eq.~\ref{eq:#1}}
\newcommand{\refTbl}[1]{Tbl.~\ref{tbl:#1}}

\newcommand{\Pprojected}{K} % 2D keypoints
\newcommand{\Proot}{P} % 3D root centeres
\newcommand{\PG}{P^{[G]}} % 3D global...
\newcommand{\PF}{P_\text{fused}}
\newcommand{\PS}{P_\text{sum}}
\newcommand{\PD}{P_\text{deep}}
\newcommand{\Ofirst}{O\vspace{-1pt}\emph{1}}
\newcommand{\Osecond}{O\vspace{-1pt}\emph{2}}
\newcommand{\dataset}{MPI-INF-3DHP}
\newcommand\blfootnote[1]{%
  \begingroup
  \renewcommand\thefootnote{}\footnote{#1}%
  \addtocounter{footnote}{-1}%
  \endgroup
}
%Table related stuff
\newcommand\T{\rule{0pt}{2.6ex}}       % Top strut
\newcommand\B{\rule[-1.2ex]{0pt}{0pt}} % Bottom strut
%%%%%%%%% TITLE
%\title{Monocular 3D Human Pose Estimation Using Improved CNN Supervision and Transfer Learning}
\makeatletter
\renewcommand\AB@affilsepx{ ~~~  \protect\Affilfont}
\makeatother
\title{Monocular 3D Human Pose Estimation In The Wild\\Using Improved CNN Supervision}

\author[1]{\vspace{-3ex}Dushyant Mehta}
\author[2]{Helge Rhodin}
\author[3]{Dan Casas}
\author[2]{Pascal Fua}
\author[1]{\\Oleksandr Sotnychenko}
\author[1]{Weipeng Xu}
\author[1]{Christian Theobalt}
\affil[1]{MPI for Informatics, Germany}
\affil[2]{EPFL, Switzerland}
\affil[3]{Universidad Rey Juan Carlos, Spain\vspace{-4ex}}
\maketitle
\renewcommand\Authands{ and }
%\thispagestyle{empty}
%%%%%%%%% ABSTRACT
\begin{abstract}
We propose a CNN-based approach for 3D human body pose estimation from single RGB images that addresses the issue of limited generalizability of models trained solely on the starkly limited publicly available 3D pose data.
Using only the existing 3D pose data and 2D pose data, we show state-of-the-art performance on established benchmarks through transfer of learned features, while also generalizing to in-the-wild scenes.
We further introduce a new training set for human body pose estimation from monocular images of real humans that has the ground truth captured with a multi-camera marker-less motion capture system. It complements existing corpora with greater diversity in pose, human appearance, clothing, occlusion, and viewpoints, and enables an increased scope of augmentation.
We also contribute a new benchmark that covers outdoor and indoor scenes, and demonstrate that our 3D pose dataset shows better in-the-wild performance than existing annotated data, which is further improved in conjunction with transfer learning from 2D pose data. 
All in all, we argue that the use of transfer learning of representations in tandem with algorithmic and data contributions is crucial for general 3D body pose estimation.
\ignore{We propose novel CNN supervision techniques, using a regularization structure while training that extends the concept of multi-level skip connections, and leverage first
and second order parent relationships along the skeletal kinematic tree to learn better representations.}
\vspace{-0.4cm}
\end{abstract}

%\input{Sections/introduction}
%!TEX root = ../article.tex
\mysection{Introduction}{sec:intro}
We present an approach to estimate the 3D articulated human body pose from a single image taken in an uncontrolled environment.\blfootnote{This work was funded by the ERC Starting Grant project CapReal (335545). Dan Casas was supported by a Marie Curie Individual Fellow grant (707326), and Helge Rhodin by the Microsoft Research Swiss JRC. We thank The Foundry for license support.}
Unlike 
marker-less 3D motion capture methods that {\it track} articulated human poses from {\it multi-view} video sequences, 
\cite{wren_pfinder_pami1997,sminchisescu_covariance_cvpr2001,starck_model_iccv2003,urtasun_monocular_cvpr2005,balan_detailed_cvpr2007,gall_optimization_ijcv2010,stoll_fast_iccv2011,chai2005performance} or use \emph{active} RGB-D cameras~\cite{shotton_poseparts_acm13,baak_posedepth_iccv11},
our approach is designed to work from a single low-cost RGB camera.

Data-driven approaches using Convolutional Neural Networks (CNNs) have shown impressive results
for 3D pose regression from monocular RGB, however, in-the-wild scenes and motions remain challenging.
Aside from the difficulty of the 3D pose estimation problem, it is further stymied by the lack of suitably large and diverse annotated 3D pose corpora.
For 2D joint detection it is feasible to obtain ground truth annotations on in-the-wild data on a large scale through crowd sourcing~\cite{sapp_modec_cvpr13,andriluka_mpii2d_cvpr14,johnson_lsp_bmvc10}, consequently leading to methods that generalize to in-the-wild scenes 
~\cite{chen_nips14,wei_cpm_cvpr16,tompson_cnn_graph_pose_nips14,toshev_deeppose_cvpr14,
pishchulin_deepcut_cvpr16,bulat_convpart_eccv16,belagiannis_recurrent_arxiv6,
newell_stacked_hourglass_eccv16,
lifshitz_deep_consensus_eccv16,
gkioxari_chained_eccv2016,hu_bottomup_cvpr16,
carreira_iterative_cvpr16}.
Some 3D pose estimation approaches take advantage of this generalizability of 2D pose estimation, and propose to lift the 2D keypoints to 3D \cite{tome_lifting_2017,yasin_dual_source_cvpr16,bogo_smpl_eccv16,wang_robust_cvpr2014,li_maximum_iccv2015,zhou_convexrelaxation_cvpr2015,zhou_sparseness_deepness_cvpr15,zhou_spatio_eccv2014,simo_single_cvpr2012,simo_joint_CVPR2013,chen_2d_match_cvpr17}. This approach however is susceptible to errors from depth ambiguity, and often requires computationally expensive iterative pose optimization.
Recent advances in direct CNN-based 3D regression show promise, utilizing different prediction space formulations \cite{tekin_structured_bmvc16,li_accv14,zhou_deep_kinematic_arxiv16,pavlakos_volumetric_cvpr17,moreno_distance_matrix_cvpr17} and incorporating additional constraints \cite{zhou_deep_kinematic_arxiv16,tekin_motion_comp_cvpr16,zhou_sparseness_deepness_cvpr15, yu_mono_heightmap_eccv16}. However, we show on a new in-the-wild benchmark that existing solutions have a low generalization to in-the-wild conditions. They are far from the accuracy seen for 2D pose prediction in terms of correctly located keypoints.

\begin{figure*}[t!]
	\centering\includegraphics*[page=2,width=\linewidth,trim={0 13.3cm 0.5cm 0},clip]{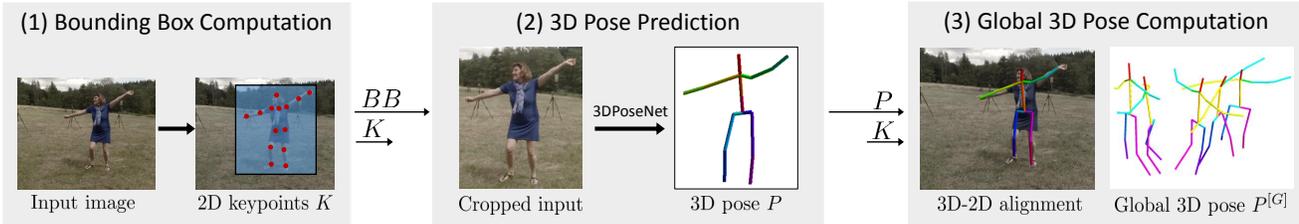}
	%	\centering\includegraphics*[page=2,width=\linewidth,trim={0 13.3cm 0cm 0},clip]{\figurePath/Overview/overview.pdf}
	\caption{We infer 3D pose from single image in three stages: (1) extraction of the actor bounding box from 2D detections; (2) direct CNN-based 3D pose regression; and (3) global root position computation in original footage by aligning 3D to 2D pose.}
	\label{fig:Oveview_prediction}
    \vspace{-0.35cm}
\end{figure*}
Existing 3D pose datasets use marker-based motion capture, MoCap, for 3D annotation \cite{ionescu_human36_pami14,sigal_humaneva_ijcv10}, which restricts recording to skin-tight clothing,
or markerless systems in a dome of hundreds of cameras \cite{joo_panoptic_iccv2015}, which enables diverse clothing but requires an expensive studio setup.
Synthetic data can be generated by retargeting MoCap sequences to 3D avatars \cite{chen_synth_data_3dv16}, however the results lack realism, and learning based methods pick up on the peculiarities of the rendering leading to poor generalization to real images.

Our contributions towards accurate in-the-wild pose estimation are twofold.
First, in Section \ref{sec:transfer-learning}, we explore the use of transfer learning to leverage the highly relevant mid- and high-level features learned on the readily available in-the-wild 2D pose datasets \cite{andriluka_mpii2d_cvpr14,johnson_lspet_cvpr11} in conjunction with the existing annotated 3D pose datasets.
%In addition to the aforementioned architectural improvements.
Our experimentally validated mechanism of feature transfer shows better accuracy and generalizability compared to na\"ive weight initialization from 2D pose estimation networks and domain adaptation based approaches. With this we show previously unseen levels of accuracy on established benchmarks, as well as generalizability to in-the-wild scenes, with only the existing 3D pose datasets. 

Second, in Section \ref{sec:mpii3d}, we introduce the new 
\dataset~dataset \footnote{MPI-INF-3DHP dataset available at \url{gvv.mpi-inf.mpg.de/3dhp-dataset}} real humans with ground truth 3D annotations from a state-of-the-art markerless motion capture system.
It complements existing datasets with everyday clothing appearance,  a large range of motions, interactions with objects, and more varied camera viewpoints. 
The data capture approach eases appearance augmentation to extend the captured variability, complemented with improvements to existing augmentation methods for enhanced foreground texture variation. This gives a further significant boost to the accuracy and generalizability of the learned models. 

\change{The data-side supervision contributions are complemented by CNN architectural supervision contributions in Section~\ref{sec:3d-regression}, which are orthogonal to in-the-wild performance improvements.}

Furthermore, we introduce a new test set, including sequences outdoors with accurate annotation, on which we demonstrate the generalization capability of the proposed method and validate the value of our new dataset.

The components of our method are thoroughly evaluated on existing test datasets, demonstrating both state-of-the-art results in controlled settings and, more importantly, improvements over existing solutions for in-the-wild sequences thanks to the better generalization of the proposed techniques.

%
%!TEX root = ../article.tex
\mysection{Related Work}{sec:related-work}
\change{There has been much work on learning- and model-based approaches for human body pose estimation} from monocular images, with much of the recent progress coming through CNN based approaches. We review the most relevant approaches, and discuss their relation with our work.

%%%%%%%%%%%%%%%%%%%%%%%%%%%%%%%%
%
\noindent\textbf{3D pose from 2D estimates.}
Deep CNN architectures have dramatically improved 2D pose estimation 
\cite{jain_modeep_accv14,newell_stacked_hourglass_eccv16}, \change{with even real-time solutions}
\cite{wei_cpm_cvpr16}.
Graphical models \cite{felzenszwalb_pictorial_ijcv05,agarwal_recovering_pami06} \change{continue to find use in }
modeling multi-person relations \cite{pishchulin_deepcut_cvpr16}.
3D pose can be inferred from 2D pose through geometric and statistical priors~\cite{mori_contexts_pami06,taylor_articulated_cvpr00}.
\change{Optimization of the projection of a 3D human model to the 2D predictions is computationally expensive and ambiguous, but the ambiguity can be addressed through pose priors and it further allows incorporation of various constraints such as }
inter-penetration constraints \cite{bogo_smpl_eccv16},
sparsity assumptions \cite{wang_robust_cvpr2014,zhou_convexrelaxation_cvpr2015,zhou_sparse_arXiv2015},
joint limits \cite{elhayek_convmocap_TPAMI2016,akhter_pose_conditioned_cvpr15}, 
and temporal constraints \cite{rhodin_general_eccv16}.
%
%An alternative to iterative optimization is sampling. 
Simo-Serra \etal \cite{simo_single_cvpr2012} sample noisy 2D predictions to ambiguous 3D shapes, which they disambiguate using kinematic constraints, and improve discriminative 2D detection from likely 3D samples \cite{simo_joint_CVPR2013}.
Li \etal look up the nearest neighbours in a learned joint embedding of human images and 3D poses~\cite{li_maximum_iccv2015} to estimate 3D pose from an image.
We choose to use the geometric relations between the predicted 2D and 3D skeleton pose to infer the global subject position.
%%%%%%%%%%%%%%%%%%%%%%%%%%%%%%%%
% End-to-end 3D pose prediction
\noindent\textbf{Estimating 3D pose directly.}
Additional image information, e.g.\ on the front-back orientation of limbs, can be exploited by regressing 3D pose directly from the input image \cite{tekin_structured_bmvc16,li_accv14,zhou_deep_kinematic_arxiv16,ionescu_iterated_cvpr14}.
Deep CNNs achieve state-of-the-art results \cite{zhou_deep_kinematic_arxiv16,tekin_fusion_arxiv16,pavlakos_volumetric_cvpr17}.
While CNNs dominate, regression forests have also been used
to derive 3D \emph{posebit descriptors} efficiently
%, answering questions such as "is the left leg in-front of the right leg", to query 3D pose with corresponding annotation 
\cite{pons_posebits_cvpr14}. 
The input and output representations are important too. 
To localize the person, the input image is commonly cropped to the bounding box of the subject before 3D pose estimation \cite{ionescu_iterated_cvpr14}.
Video input provides temporal cues, which translate to increased accuracy \cite{tekin_motion_comp_cvpr16,zhou_sparseness_deepness_cvpr15}. The downside of conditioning on motion is the increased input dimensionality, and requires motion databases with sufficient motion variation, which are even harder to capture than pose data sets.
In controlled conditions, fixed camera placement provides additional height cues \cite{yu_mono_heightmap_eccv16}.
Since monocular reconstruction is inherently scale-ambiguous, 
3D joint positions relative to the pelvis, with normalized subject height  are widely used as the output. 
To explicitly encode dependencies between joints,
Tekin \etal~\cite{tekin_structured_bmvc16} regressing to a high-dimensional pose representation, learned by an auto encoder.
Li \etal~\cite{li_accv14} report that predicting positions relative to the parent joint of the skeleton improves performance,
but we show that a pose-dependent combination of absolute and relative positions leads to further improvements.
Zhou \etal \cite{zhou_deep_kinematic_arxiv16} regress joint angles of a skeleton from single images, using a kinematic model.

%\paragraph{Transfer learning in pose estimation}
\noindent\textbf{Addressing the scarcity and limited appearance variability of datasets.}
Learning-based methods require large annotated dataset corpora. 3D annotation \cite{ionescu_iterated_cvpr14} is harder to obtain than 2D pose annotation.
Some approaches treat 3D pose as a hidden variable, and use pose priors and projection to 2D to guide the training~\cite{brau_annotations_3dv16,yasin_dual_source_cvpr16}.
Rogez \etal render mosaics of in-the-wild human pose images using projected mocap data \cite{rogez_mocap_nips16}. 
Chen \etal \cite{chen_synth_data_3dv16} render textured rigged human models, but still require domain adaptation to in-the-wild images for generalization.
Other approaches use the estimated 2D pose to look up a suitable 3D pose from a dictionary \cite{chen_2d_match_cvpr17}, or use the ground truth 2D pose based dictionary lookup to create 3D annotations for in-the-wild 2D pose data \cite{rogez_lcr_cvpr17}, but neither address the 2D to 3D ambiguity.
Our new dataset complements the existing datasets, through extensive appearance and pose variation, by using marker-less annotation and provides an increased scope for augmentation.

\emph{Transfer Learning} \cite{pan_survey_kde2010} is commonly used in computer vision to leverage features and representations learned on one task to offset data scarcity for a related task.
\ignore{in case there is no data imbalance.}Low and/or mid-level CNN features can be shared also among unrelated tasks \cite{sharif_off_the_shelf_cvprw14,yosinski_transferable_nips14}. Pretraining on ImageNet~\cite{imagenet_ijcv2015} is commonly used for weight initialization~\cite{insafutdinov_deepercut_eccv16,tekin_fusion_arxiv16} in CNNs.
We explore different ways of using the low and mid-level features learned on in-the-wild 2D pose datasets for further improving the generalization of 3D pose prediction models. 

%
%!TEX root = ../article.tex
\mysection{CNN-based 3D Pose Estimation}{sec:3d-pose}
We start by introducing the network architecture, utilized input and output domains, and notation.
While the particularities of our architecture are explained in Section \ref{sec:3d-regression}, 
our main contributions towards in-the-wild conditions are covered in sections \ref{sec:transfer-learning} and \ref{sec:mpii3d}.
\myfigure{Skeleton_Parent}{3D pose, represented as a vector of 3D joint positions, is expressed variously as 1) $\Proot$: relative to the root (joint \#15), 2) \Ofirst~(blue): relative to first order and, 3) \Osecond~(orange): relative to second order parents in the kinematic skeleton hierarchy.
}{0.35}

\begin{figure*}[t!]
	\centering\includegraphics*[page=1,width=1\linewidth,trim={0 6.5cm 3.6cm 0},clip]{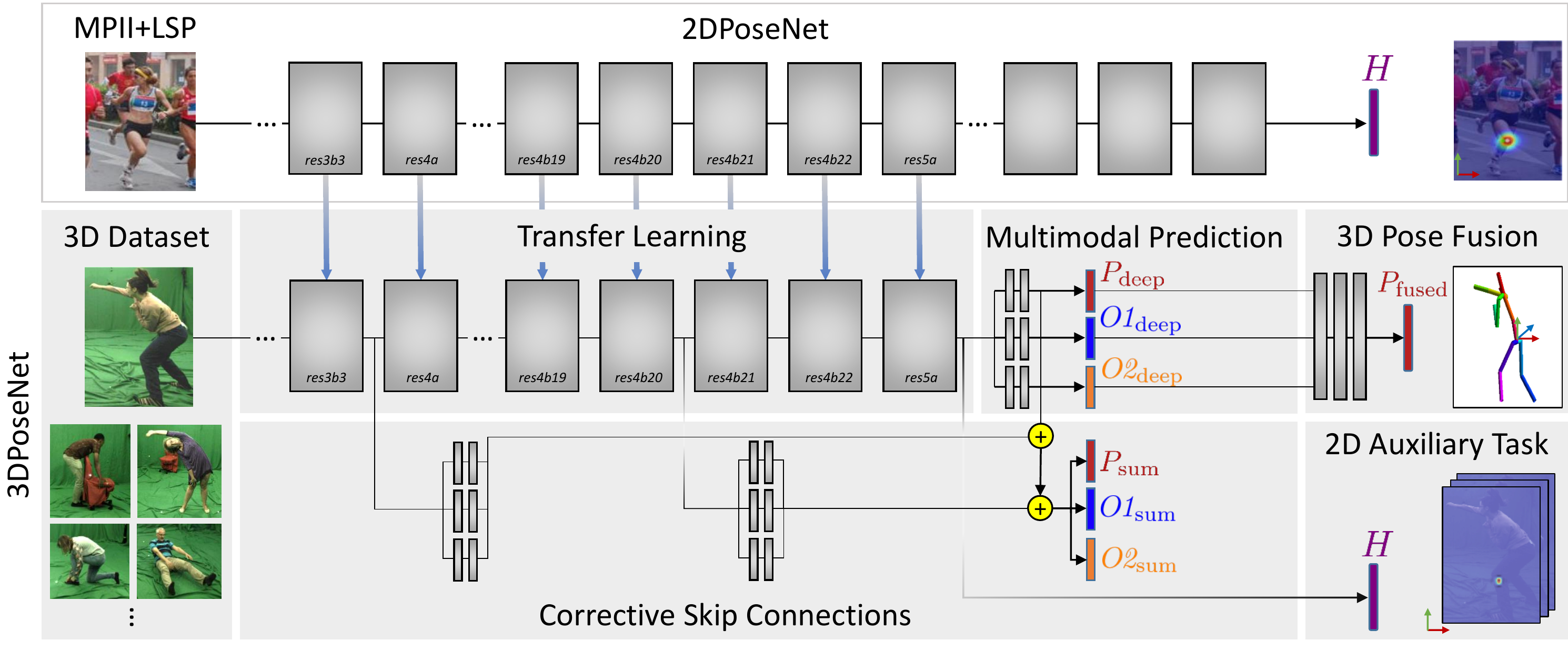}
	\caption{3D pose Training overview. The main components are 1) regularization through corrective skip connections, and 2D pose prediction as auxiliary task, 2)~Multi-modal 3D pose prediction and fusion, 3) a new marker-less 3D pose database with appearance augmentation, and 4)~Transfer learning from features learned for 2D pose estimation.}
	\label{fig:Oveview_training}
    \vspace{-0.4cm}
\end{figure*}
Given an RGB image, we estimate the global 3D human pose $\PG$ in the camera coordinate system. We estimate the global positions of the joints of the skeleton depicted in Figure \ref{fig:Skeleton_Parent}, accounting for the camera viewpoint, which goes beyond only estimating in a root-centered (pelvis) coordinate system, as is common in many previous works. 
Our algorithm consists of three steps, as illustrated in Figure \ref{fig:Oveview_prediction}.
{\bf(1)} the subject is localized in the frame with a 2D bounding box $BB$, computed from 2D joint heatmaps $H$, obtained with a CNN we call \emph{2DPoseNet};
{\bf(2)} the root-centered 3D pose $\Proot$ is regressed from the $BB$-cropped input with a second CNN termed \emph{3DPoseNet}; 
and {\bf(3)} global 3D pose coordinates $\PG$ and perspective correction are computed in closed form using 3D pose $\Proot$, 2D joint locations $\Pprojected$ %(extracted from $H$) 
and known camera calibration.  

\mysubsection{Bounding Box and 2D Pose Computation}{sec:2d-preprocessing}
We use our \textit{2DPoseNet} to produce 2D joint location heatmaps $H$.
The heat map maxima provide the most likely 2D joint locations $\Pprojected$ which can \change{also act as a stand-in person bounding-box $BB$ detector.}
See Figure \ref{fig:Oveview_prediction}.
The 2D joint locations $\Pprojected$ are further used for global pose estimation in Section \ref{sec:3d-postprocessing}. \change{In case of an alternative $BB$ detector, $\Pprojected$ comes from \textit{3DPoseNet}. See \textit{2D Auxiliary Task} in Figure \ref{fig:Oveview_training}}.

Our \textit{2DPoseNet} is fully convolutional and is trained on MPII~\cite{andriluka_mpii2d_cvpr14} and LSP~\cite{johnson_lspet_cvpr11,johnson_lsp_bmvc10} datasets.\ignore{, with images resized to $368\times368$ px}
We use a CNN structure based on Resnet-101~\cite{he_resnet_cvpr2016}, up to the filter banks at level 4.
Striding is removed at level 5, and features in the \textit{res5a} block are halved and identity skip connections removed from \textit{res5b} and \emph{res5c}.\ignore{Since we need heatmaps as output, striding is removed at level 5. Additionally, we remove the identity skip connections at level 5 and use fewer features per layer.} \ignore{The heatmaps are upsampled $\times2$ to be 1/8\textsuperscript{th} the width and height of the input.}
For specifics of the network architecture and the training scheme, refer to the supplementary document. 
\subsection{3D Pose Regression}\label{sec:3d-regression}
The 3D pose CNN, termed \textit{3DPoseNet}, is used to regress root-centered 3D pose $\Proot$ from a cropped RGB image, and makes use of new CNN supervision techniques.
Figure \ref{fig:Oveview_training} depicts the main components of the method, detailed in the following sections.

\noindent\textbf{Network} %\label{sec:network}
The base network derives from Resnet-101 as well, and is identical to \textit{2DPoseNet} up to \texttt{res5a}. We remove the remaining layers from level 5. A 3D prediction stub $\mathcal{S}$ comprised of a convolution layer ($k_{5\times5},s_{2}$) with 128 features and a final fully-connected layer that outputs the 3D joint locations is added on top. Additionally we predict 2D heatmaps $H$ as an auxiliary task after \texttt{res5a} and, use intermediate supervision with pose $\Proot$ at \texttt{res3b3} and \texttt{res4b22}. Refer to the supplementary for specifics of the loss weights for the intermediate and auxiliary tasks. 

\subsubsection{Multi-level Corrective Skip Connections} \label{sec:skip-connections}
We additionally use a skip connection scheme as a training-time regularization architecture. We add skip connections from \texttt{res3b3} and \texttt{res4b20} to the main prediction $\PD$, leading to $\PS$. In contrast to vanilla skip-connections~\cite{long_fcn_CVPR2015}, we compare both $\PS$ and $\PD$ to the ground truth, and remove the skip connections after training. 
We show the improvements due to this approach Section \ref{results}.

\subsubsection{Multi-modal Pose Fusion} \label{sec:multi-modal-pose-fusion}
Formulating joint location prediction relative to a single local or global location is not always optimal. Existing literature \cite{li_accv14} has observed that predicting joint locations relative to their direct kinematic parents (Order 1 parents) improves performance.
Our experiments reveal that to not universally hold true. We find that depending on the pose and the visibility of the joints in the input image, the optimal relative joint for each joint's location prediction differs. Hence, we use joint locations $\Proot$ relative to the root, $\Ofirst$ relative to Order 1 parents and $\Osecond$ relative to Order 2 parents along the kinematic tree as the \emph{three modes} of prediction, see Figure \ref{fig:Skeleton_Parent}, and fuse them with fully-connected layers. 

For the joint set we consider, the kinematic relationships chosen suffice, as it puts at least one reference joint for each joint in the relatively low entropy torso \cite{lehrman_bayesian_pose_iccv13}. We use three identical 3D prediction stubs attached to \texttt{res5a} for predicting the pose as $P$, $\Ofirst$ and $\Osecond$, and for each we use corrective skip connections. These predictions are fed into a smaller network with three fully connected layers, to implicitly determine and fuse the better constraints per joint into the final prediction $\PF$. The network has the flexibility to emphasize different combinations of constraints depending on the pose. 
This can be viewed as intermediate supervision with auxiliary tasks, yet the separate streams for predicting each mode individually are key to its efficacy.

\subsection{Global Pose Computation}\label{sec:3d-postprocessing}
The bounding box cropping normalizes subject size and position, which frees 3D pose regression from having to localize the person in scale and image space, but loses global pose information.
We propose a lightweight and efficient way to reconstruct
the global 3D pose $\PG = \left(R | T\right) \PF$ from pelvis-centered pose $\PF$, camera intrinsics, and $\Pprojected$.

\noindent\textbf{Perspective correction.~}
The bounding box cropping can be interpreted as using a virtual camera, 
rotated towards the crop center and its field of view covering the crop area.
Since the \textit{3DPoseNet} only `sees' the cropped input, its predictions live in this rotated view, leading to a consistent orientation error in $\PF$.
%, see Table \ref{tbl::HumanEva}.
To compensate, we compute rotation $R$ that rotates the virtual camera to the original view.
 
\noindent\textbf{3D localization.~}
We seek the global translation $T$ that aligns $\PF$ and $\Pprojected$ under perspective projection.
We assume weak perspective projection, $\Pi$, and solve the linear least squares equation $\sum_i \lVert \Pprojected^i - \Pi (T + \PF^i) \rVert^2$, where $i$ indexes the joints.
This assumption yields global position
\begin{equation}
T = 
\frac{\sqrt{\sum_i \lVert\Proot^i_{[xy]} - \bar{\Proot}_{[xy]}\rVert^2 } }{\sqrt{\sum_i \lVert\Pprojected^i - \bar{\Pprojected}\rVert^2} }
\begin{pmatrix}\bar{K}_{[x]}\\ \bar{K}_{[y]}\\f\end{pmatrix} 
- \begin{pmatrix}\bar{P}_{[x]}\\ \bar{P}_{[y]}\\0\end{pmatrix}
\text{,}
\label{eq:globalPosition}
\end{equation}
in terms of distances to the 3D mean $\bar{\Proot}$ and 2D mean $\bar{\Pprojected}$ over all joints. $\Proot_{[xy]}$ is the $x,y$ part of $\PF$ and single subscripts indicate the respective elements.
Please see the supplemental document for the derivation and evaluation.

Our solution can be considered a generalization of \emph{procrustes analysis} for projective alignment. 
Note that this is different to
\emph{perspective-n-point} 6DOF rigid pose estimation \cite{lepetit_monocular_book2005},
 structure-from-motion, and from the convex approach of Zhou \etal \cite{zhou_convexrelaxation_cvpr2015}, which require iterative optimization.

\section{Transfer Learning} \label{sec:transfer-learning}
We use the features learned with Resnet-101 from ImageNet \cite{imagenet_ijcv2015} to initialize both \emph{2DPoseNet} and \emph{3DPoseNet}, as common for many vision tasks.
While this affords a faster convergence while training, there remains room for improved generalization beyond the gains from potential supervision and dataset contributions.  
Due to the similarity of the tasks, features learned for 2D pose estimation on in-the-wild MPII and LSP training sets can be transferred to 3D pose estimation. We explore different variants of the, thus far, un-utilized method of improving generalization by transferring weights from \emph{2DPoseNet} to \emph{3DPoseNet}.

%%!TEX root = ../article.tex
\begin{table}[]
\centering
\caption{Evaluation of the mechanisms of transfer learning from \emph{2DPoseNet} to \emph{3DPoseNet} that were explored in the context of the \emph{Base} network. The table compares the effect of various learning rate multiplier combinations for different parts of the network. For network details, refer to Section \ref{sec:3d-regression}. Human3.6m, Subjects 1,5,6,7,8 used for training, and every $64^{th}$ frame of 9,11 used for testing. * = weights randomly initialized}
\label{tbl:s9_11_tx_method}
%\resizebox{1.8\columnwidth}{!}{
\resizebox{0.93\columnwidth}{!}{
\begin{tabular}{@{}ccc|c@{}}
\multicolumn{3}{c|}{\textbf{Learning Rate Multiplier}}                      &  \\ \cline{1-3}
\multicolumn{1}{c|}{{up to \texttt{res4b22}}\T} & \multicolumn{1}{c|}{{\texttt{res5a}}} & \multicolumn{1}{c|}{{3D Stub $\mathcal{S}$}}                  & \textbf{Total MPJPE (mm)} \\ \hline \hline
\multicolumn{1}{c|}{$1$\T} & \multicolumn{1}{c|}{$1$} & \multicolumn{1}{c|}{1*}    &  118.7\\ \hline
\multicolumn{1}{c|}{$1/10$\T} & \multicolumn{1}{c|}{$1/10$} & \multicolumn{1}{c|}{1*}    & 84.6 \\ \hline
\multicolumn{1}{c|}{$1/1000$\T} & \multicolumn{1}{c|}{$1/1000$} & \multicolumn{1}{c|}{1*}   & 89.2  \\ \hline
\multicolumn{1}{c|}{$1/10$\T} & \multicolumn{1}{c|}{$1$} & \multicolumn{1}{c|}{1*}   & 90.7\\ \hline
\multicolumn{1}{c|}{$1/1000$\T} & \multicolumn{1}{c|}{$1$} & \multicolumn{1}{c|}{1*}    & \textbf{80.7}  \\ \hline
%\multicolumn{1}{c|}{$1/1000$} & \multicolumn{1}{c|}{$1$*} & \multicolumn{1}{c|}{1*}    &  \\ \hline
\end{tabular}
}
\vspace{-0.5cm}
\end{table}

A na\"ive initialization of the weights of \emph{3DPoseNet} is inadequate, and there is a tradeoff to be made between the preservation of transferred features and learning new pertinent features. We achieve this through a learning rate discrepancy between the transferred layers and the new layers. We experimentally determine the mechanism for this transfer of features through validation. Table \ref{tbl:s9_11_tx_method} shows the evaluated mechanisms for transfer from \emph{2DPoseNet}. Based on the experiments, we choose to scale down the learning rate of the layers till \texttt{res4b22} by a factor of 1000. Through similar experiments for the transfer of ImageNet features, we choose to scale down the learning rate of layers till \texttt{res4b22} by 10.

The same approach can be applied to other network architectures, and our experiments on the learning rate discrepancy serve as a sound starting point for the determination of the transfer learning mechanism. \change{Unlike jointly training with annotated 2D and 3D pose datasets, this approach has the advantage of not requiring the 2D annotations to be consistent between the two datasets, and one can simply use off-the-shelf trained 2D pose networks. }
In Section \ref{results} we show that our approach outperforms domain adaptation, see Table \ref{tbl:our_test_txlearn}, first row.
Additionally, Table \ref{tbl:s9_11_tx_method} validates that the common fine-tuning of the fully-connected layers (third row) and fine-tuning of the complete network (first row) is much less effective then the proposed scheme.

\section{\dataset: Human Pose Dataset }\label{sec:mpii3d}
We propose a new dataset captured in a multi-camera studio with ground truth from commercial marker-less motion capture~\cite{captury}. 
No special suits and markers are needed, allowing the capture of motions wearing everyday apparel, including loose clothing. In contrast to existing datasets, we record in green screen studio to allow automatic segmentation and augmentation.
We recorded 8 actors (4m+4f), performing 8 activity sets each, ranging from walking and sitting to complex exercise poses and dynamic actions, covering more pose classes than Human3.6m. Each activity set spans roughtly one minute. Each actor features 2 sets of clothing split across the activity sets. One clothing set is \emph{casual everyday apparel}, and the other is \emph{plain-colored} to allow augmentation.

We cover a wide range of viewpoints, with five cameras mounted at chest height with a roughly 15$^{\circ}$ elevation variation similar to the camera orientation jitter in other datasets \cite{chen_synth_data_3dv16}. Another five cameras are mounted higher and angled down 45$^{\circ}$, three more have a top down view, and one camera is at knee height angled up.
Overall, from all 14 cameras, we capture $>$1.3M frames, 500k of which are from the five chest high cameras. We make available both true 3D annotations, and a skeleton compatible with the ``universal'' skeleton of Human3.6m .   

\begin{figure}
	\centering
	\includegraphics[trim={10cm 8cm 30cm 30cm },clip,width=0.241\columnwidth]{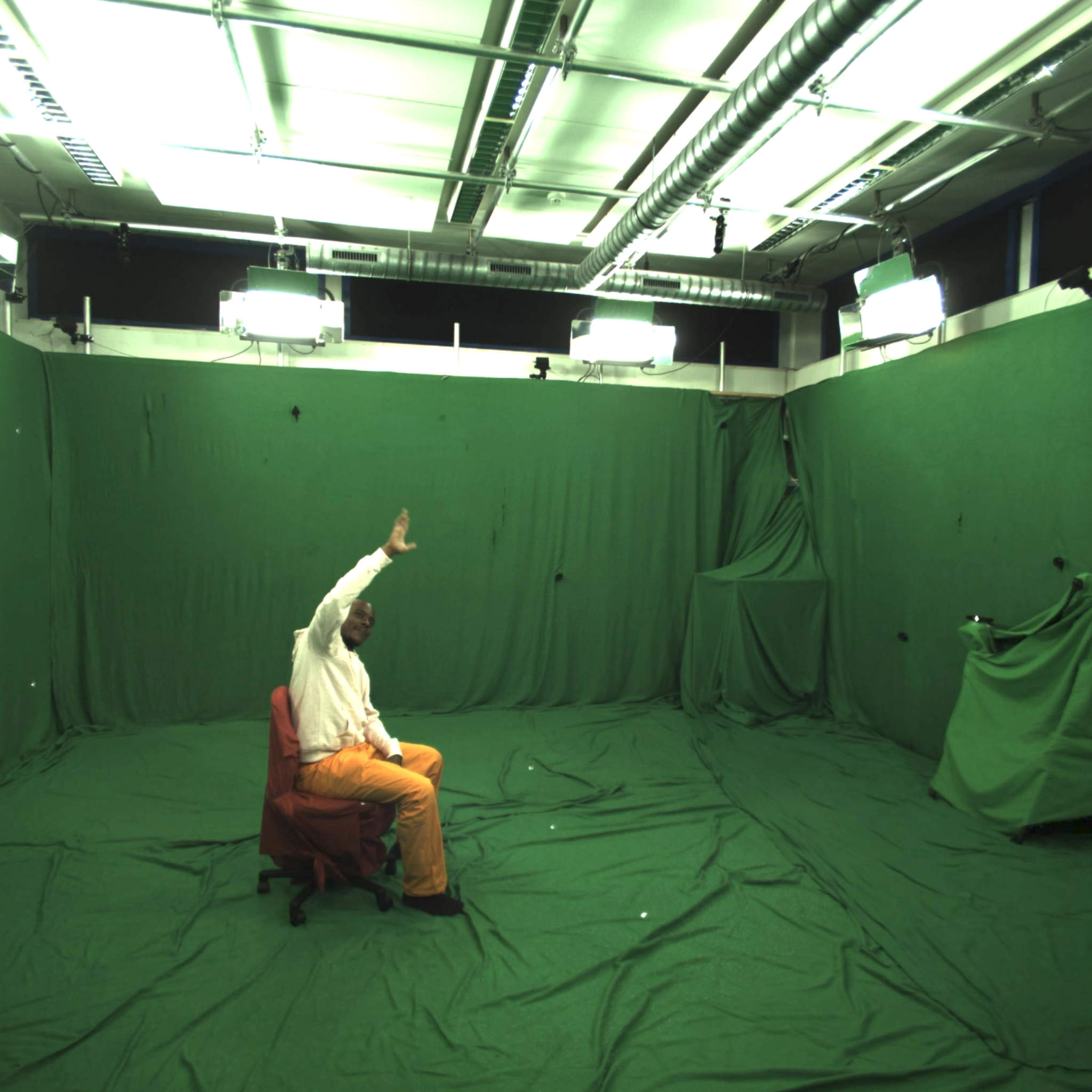}
	\includegraphics[trim={10cm 8cm 30cm 30cm },clip,width=0.241\columnwidth]{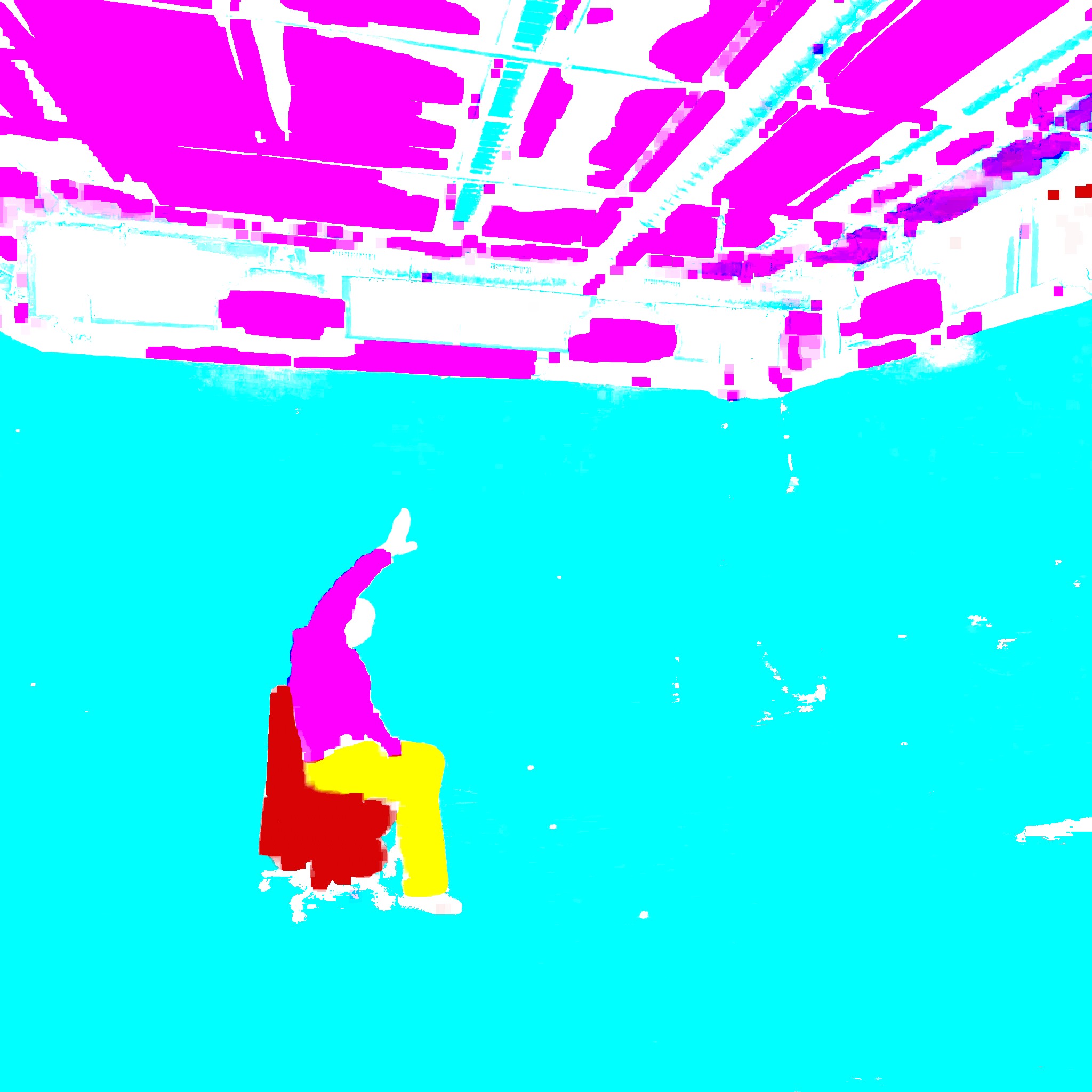}
	\includegraphics[trim={10cm 8cm 30cm 30cm },clip,width=0.241\columnwidth]{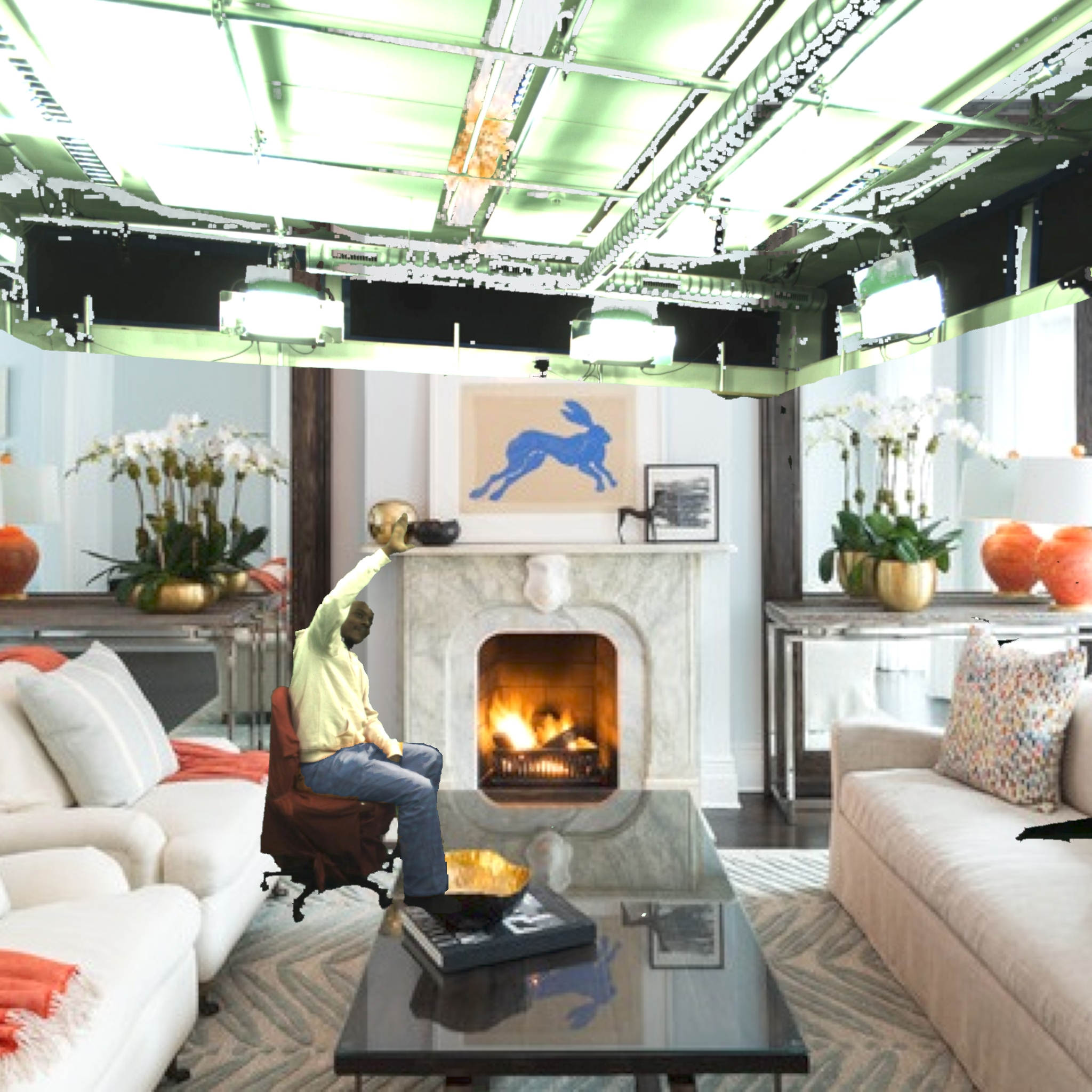}
	\includegraphics[trim={10cm 8cm 30cm 30cm },clip,width=0.241\columnwidth]{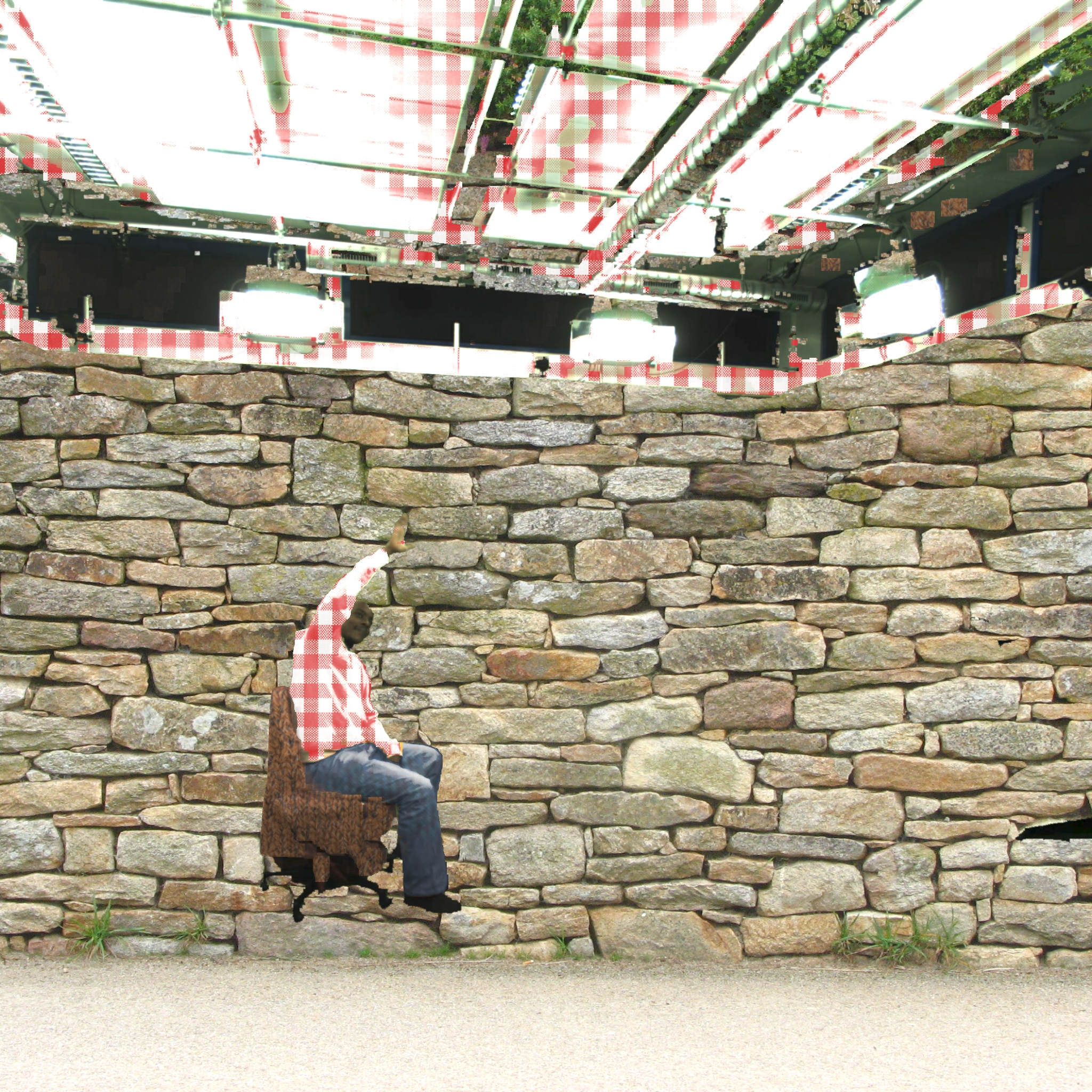}
	\caption{\dataset~dataset. We capture actors using a markerless multi-camera in a green screen studio (left), compute masks for different regions (center left) and augment the captured footage by compositing different textures to the background, chair, upper and lower body areas, independently (center right and right).}	 \label{fig:MPI-3DHP-augmentation}
    \vspace{-0.5cm}
\end{figure}
\noindent\textbf{Dataset Augmentation.~}
%Despite having alleviated the restriction to body fitting clothes, and having 
Although our dataset has more clothing variation than other datasets, the appearance variation is still not comparable to in-the-wild images. There have been several approaches proposed to enhance appearance variation.
Pishchulin \etal warp human size in images with a parametric body model~\cite{pishchulin_reshape_cvpr12}.
Images can be used to augment background of recorded footage \cite{rhodin_egocap_SIGGRAPHAsia2016, chen_synth_data_3dv16, ionescu_human36_pami14}. Rhodin \etal \cite{rhodin_egocap_SIGGRAPHAsia2016} recolor plain-color shirts while keeping the shading details, using intrinsic image decomposition to separate reflectance and shading \cite{meka2016live}.

We provide chroma-key masks for the background, a chair/sofa in the scene, as well as upper and lower body segmentation for the plain-colored clothing sets. This provides an increased scope for foreground and background augmentation, in contrast to the marker-less recordings of Joo \etal \cite{joo_panoptic_iccv2015}.
For background augmentation, we use images sampled from the internet. For foreground augmentation, we use a simplified intrinsic decomposition. 
Since for plain colored clothing the intensity variation is solely due to shading, we use the average pixel intensity as a surrogate for the shading component. We composite cloth like textures with the pixel intensity of the upper body, lower body and chair marks independently, for a photo-realistic result.  
Figure \ref{fig:MPI-3DHP-augmentation} shows example captured and augmented frames. 

\begin{figure}
	\centering
	\includegraphics[width=\columnwidth]{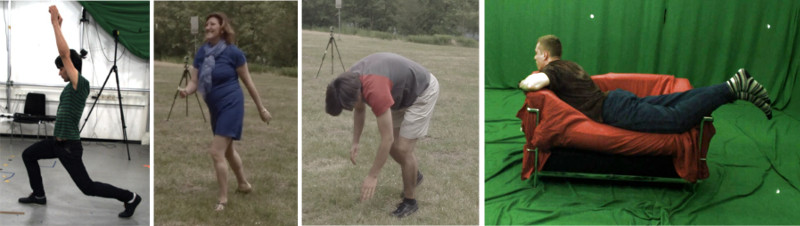}
	\caption{Representative frames from \dataset~test set. We cover a variety of subjects with a diverse set of clothing and poses in 3 different settings: studio with green screen (right); studio without green screen (left); and outdoors (center).}
	\label{fig:MPI-INF-3DHP-test}
    \vspace{-0.4cm}
\end{figure}

%!TEX root = ../article.tex
\begin{table*}[]
\centering
\caption{Activity-wise results (MPJPE in mm) on Human3.6m \cite{ionescu_human36_pami14}. Adding our model components one-by-one on top of the \emph{Base} network shows successive improvement of the total accuracy. Significant relative improvements greater than 5mm are underlined. Models are trained on Human3.6m, with network weights initialized from ImageNet, unless specified otherwise. 
The version marked with \dataset\ is trained with Human3.6m and \dataset. Evaluation with all 17 joints, on every 64\textsuperscript{th} frame, without rescaling to a person specific skeleton.
%using GT Bounding boxes for crops.
}

\label{tbl:s9_11_w_imagenet}
\setlength \tabcolsep{1.5pt}
\resizebox{1\linewidth}{!}{
\begin{tabular}{@{}|l|c|c|c|c|c|c|c|c|c|c|c|c|c|c|c||r|@{}}
\hline
\multicolumn{1}{|l|}{}   & \multirow{2}{*}{\textbf{Direct}}  &    \multirow{2}{*}{\textbf{Discuss}}           &    \multirow{2}{*}{\textbf{Eating}}      &       \multirow{2}{*}{\textbf{Greet}}                &        \multirow{2}{*}{\textbf{Phone}}  &     \multirow{2}{*}{\textbf{Posing}}      &    \multirow{2}{*}{\textbf{Purch.}}   &     \multirow{2}{*}{\textbf{Sitting}}  & \textbf{Sit} \T &    \multirow{2}{*}{\textbf{Smoke}}  & \textbf{Take} &   \multirow{2}{*}{\textbf{Wait}} & \multirow{2}{*}{\textbf{Walk}} & \textbf{Walk} & \multicolumn{1}{l||}{\textbf{Walk}} &  \multicolumn{1}{|c|}{\multirow{2}{*}{\textbf{Total}}}  \\
\multicolumn{1}{|l|}{}                      &    &  &    &  &  &    &    &      & \textbf{Down} &    & \textbf{Photo} &  &  & \textbf{Dog} & \multicolumn{1}{l||}{\textbf{Pair}} & \multicolumn{1}{|c|}{} \\ \hline \hline
\multicolumn{1}{|l|}{\T Base + Regular Skip}                                                                       & 113.34       & 112.26  & 97.40        & 110.50  & 108.63  & 112.09   & 105.67                         & 125.97 & 173.41       & 109.34  & 120.87       & 107.75  & 97.30   & 126.05   & \multicolumn{1}{r|}{117.45}    & 115.29     \\ \hline 
\multicolumn{1}{|l|}{\T Base}                                                                      & 98.98        & 100.14  & 86.07        & 101.83  & 101.34  & 96.74    & 94.89                          & 125.28 & 158.31       & 100.21  & 112.49       & 99.57   & 83.39   & 109.61   & \multicolumn{1}{r|}{95.79}     & 104.32     \\ \hline
%\multicolumn{1}{r|}{+ Fusion}                                                               & 98.20        & 99.09   & 84.84        & 100.60  & 99.25   & 95.31    & 92.38                          & 122.46   & 151.56       & 98.09   & 110.77       & 98.64   & 81.43   & 107.69   & \multicolumn{1}{r|}{93.85}     & 102.33   \\ \hline
\multicolumn{1}{|l|}{+ Corr. Skip \T }                                                                     & \underline{92.57}        & 99.08   & 85.46        & \underline{95.43}   & 96.93   & \underline{89.56}    & 95.67                          & 123.54  & 160.98       & 97.13   & \underline{107.56}       & \underline{93.86}   & \underline{76.99}   & 110.93   & \multicolumn{1}{r|}{\underline{88.73}}     & 101.09    \\ \hline
\multicolumn{1}{|l|}{+ Fusion \T}                                                                & {93.80}        & {99.17}   & {84.73}        & {95.60}   & {94.48}   & {89.40}    & {93.15}                          & {119.94}   & \underline{154.61}       & {95.94}   & {106.09}       & {94.13}   & {77.25}   & {108.82}   & \multicolumn{1}{r|}{{87.38}}     & {99.79}\\  \hline
%\multicolumn{1}{r|}{+ Fusion \textsuperscript{B,64,S}}                                                                & 52.57 & 64.09 & 55.19 & 62.23 & 71.59 & 52.75 & 68.61 & 91.84   \\ \hline 
%\multicolumn{17}{|l|}{\textbf{\T Transfer Learning from 2DPoseNet weights}}                                                                                                                                                                              \\ \hline
\multicolumn{1}{|l|}{+ Transfer \emph{2DPoseNet} \T}                                                                & \underline{59.69}        & \underline{69.74}   & \underline{60.55}        & \underline{68.77}   & \underline{\textbf{76.36}}   & \underline{59.05}    & \underline{75.04}                          & \underline{\textbf{96.19}}   & \underline{122.92}       & \underline{70.82}   & \underline{85.42}        & \underline{68.45}   & \underline{\textbf{54.41}}   & \underline{82.03}    & \multicolumn{1}{r|}{\underline{\textbf{59.79}}}     & \underline{74.14}\\  \hline
\multicolumn{1}{|l|}{+ \dataset\T}                                                                & \textbf{57.51}        & \textbf{68.58}   & \textbf{59.56}        & {\textbf{67.34}}   & {78.06}   & \textbf{56.86}    & \underline{\textbf{69.13}}                          & {99.98}   & \underline{\textbf{117.53}}       & \textbf{69.44}   & \textbf{82.40}        & \textbf{67.96}   & {55.24}   & \underline{\textbf{76.50}}    & \multicolumn{1}{r|}{{61.40}}     & \textbf{72.88}\\  \hline
\end{tabular}
}
\vspace{-0.3cm}
\end{table*}
\noindent\textbf{Test Set.~}
We found the existing test sets for (monocular) 3D pose estimation to be restricted to limited settings due to the difficulty of obtaining ground truth labels in general scenes. 
HumanEva \cite{sigal_humaneva_ijcv10} and Human3.6m \cite{ionescu_human36_pami14} are recorded indoors and test on similar looking scenes as the training set, 
the Human3D+ \cite{chen_synth_data_3dv16} test set was recorded with sensor suits that influence appearance and lacks global alignment, and
the MARCoNI set \cite{elhayek_convmocap_TPAMI2016} is markerless through manual annotation, but shows mostly walking motions and multiple actors, which are not supported by most monocular algorithms.
We create a new test set with ground truth annotations coming from a multi-view markerless motion capture system.
It complements existing test sets with more diverse motions (standing/walking, sitting/reclining, exercise, sports (dynamic poses), on the floor, dancing/miscellaneous), camera view-point variation, larger clothing variation (\eg dress), and outdoor recordings from Robertini \etal \cite{robertini_model_3dv2016} in unconstrained environments. This makes the test set suitable for testing the generalization of various methods. See Figure \ref{fig:MPI-INF-3DHP-test} for a representative sample. We use the ``universal" skeleton for evaluation. 
\begin{figure}[t]
	\centering\includegraphics*[width=\columnwidth]{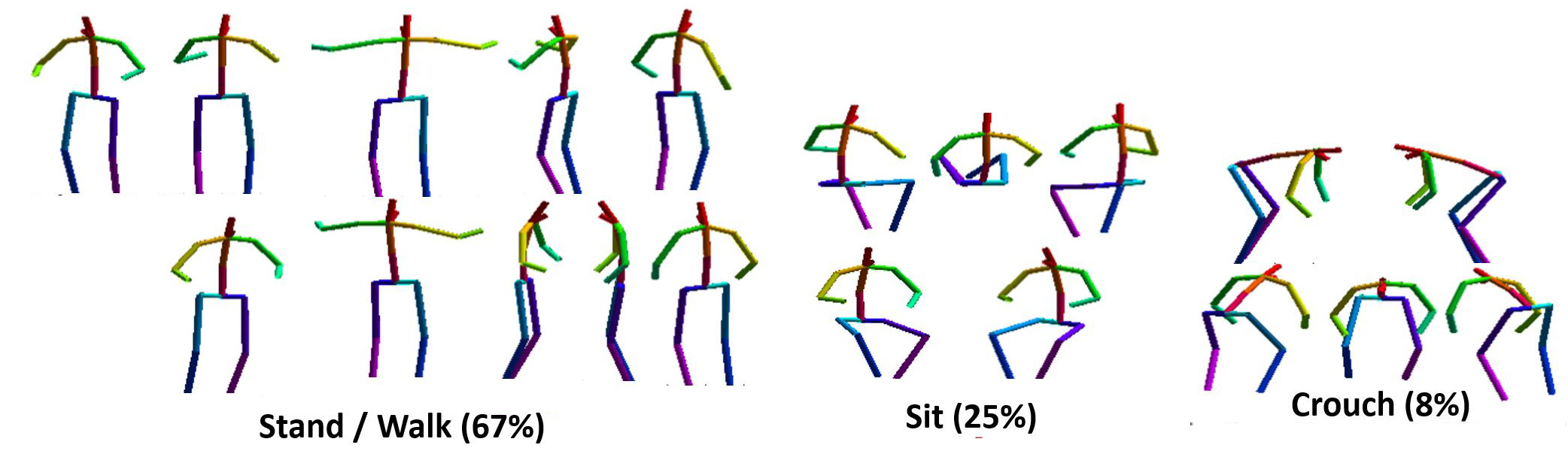}   \caption{Representative poses (centroids) of the 20 K-means pose clusters of the Human3.6m test set (subjects S9,S11), visually grouped into three broad pose classes, which are used also to perform per-class evaluation. Upright poses are dominant, with complex poses such as sitting and crouching only accounting for 25\% and 8\% of the poses respectively. Our multimodal fusion scheme significantly improves the latter two, yielding a 3.5mm improvement for Sit and 5.5mm for Crouch class.}
	\label{fig:pose_clusters}
    \vspace{-0.5cm}
\end{figure}
 
 \noindent\textbf{Alternate Metric.~}
In addition to the Mean Per Joint Position Error (MPJPE) widely used in 3D pose estimation, we concur with~\cite{ionescu_human36_pami14} and suggest a 3D extension of the Percentage of Correct Keypoints (PCK) \cite{toshev_deeppose_cvpr14,tompson_cnn_graph_pose_nips14} metric used for 2D Pose evaluation, as well as the Area Under the Curve (AUC)~\cite{insafutdinov_deepercut_eccv16} computed for a range of PCK thresholds. These metrics are more expressive and robust than MPJPE, revealing individual joint mispredictions more strongly.  
We pick a threshold of 150mm, corresponding to roughly half of head size, similar what is used in MPII 2D Pose dataset. 
We propose evaluating on the common set of joints across 2D and 3D approaches (joints 1-14 in Figure \ref{fig:Skeleton_Parent}), to ensure evaluation compatibility with existing approaches. Joints are grouped by bilateral symmetry (ankles, wrists, shoulders, etc), and can be evaluated by scene setting or activity class.

%
%!TEX root = ../article.tex
\section{Experiments and Evaluation}
\begin{figure}[]
	\centering\includegraphics*[width=0.9\columnwidth]{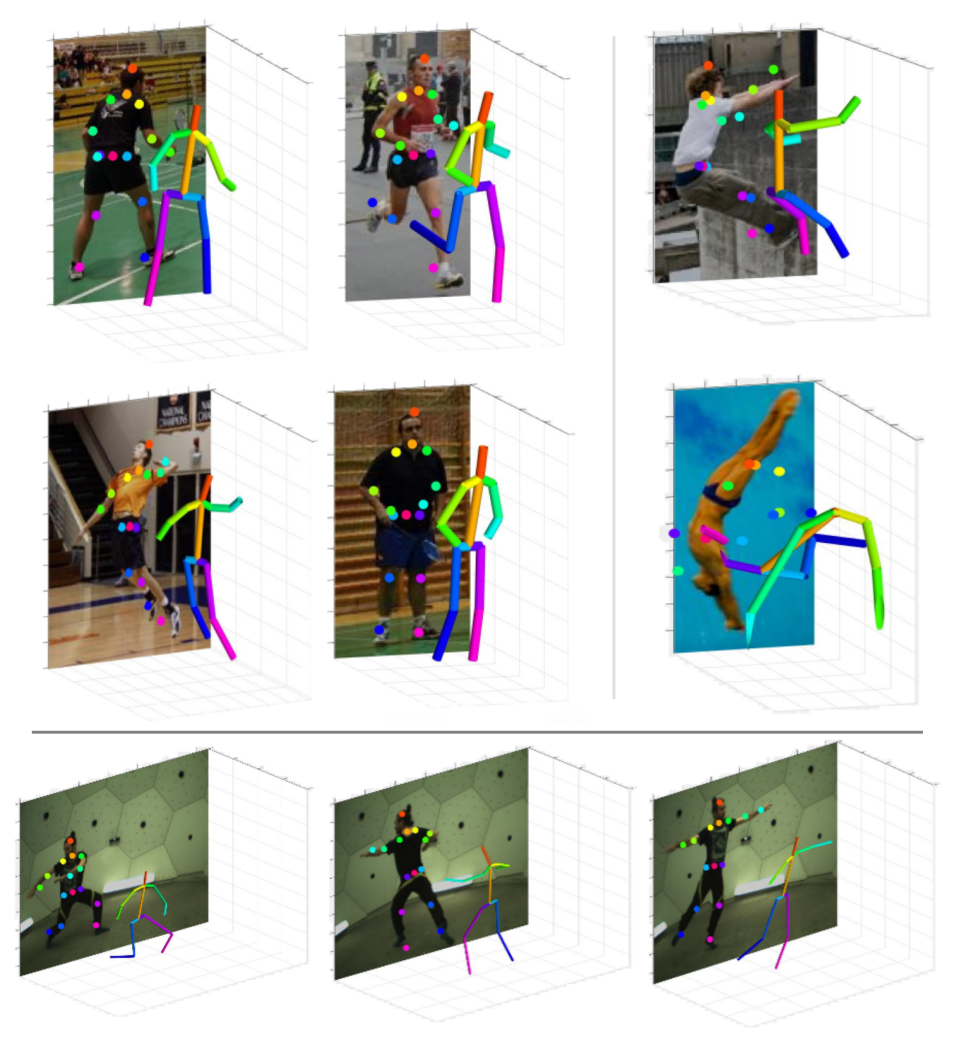}
	\caption{Qualitative evaluation on representative frames of the LSP test set. We succeed in challenging cases (left), with only few failure cases (right). The \emph{Dance1} sequence of the PanopticDataset \cite{joo_panoptic_iccv2015}, is also well reconstructed (bottom).}
	\label{fig:qualitative}
    \vspace{-0.5cm}
\end{figure}

\label{results}
We evaluate the contributions proposed in the previous sections using the standard datasets Human3.6m and HumanEva, as well as our new \dataset~test set.
Additionally, we qualitatively observe the performance on LSP \cite{johnson_lsp_bmvc10} and the CMU Panoptic \cite{joo_panoptic_iccv2015} datasets, demonstrating robustness to general scenes. Refer to Figure \ref{fig:qualitative}. Also refer to the supplementary video for global 3D pose results.

We evaluate the impact of training \emph{3DPoseNet} on Human3.6m, and unaugmented and augmented variants of \dataset, both with and without transfer learning from \emph{2DPoseNet}.
\ignore{, with  $\approx$75k frames for each after image scale augmentation at 2 scales ($0.7\times$ and $1\times$).} We only use Human3.6m compatible camera views from~\dataset~for training. Further details are in the supplemental document.

\mysubsection{Impact of Supervision Methods}{supervision_methods}

\noindent\textbf{Multi-level corrective skip connections.~} 
In Table \ref{tbl:s9_11_w_imagenet} we compare a baseline method without any skip connections, a network with vanilla skip connections, and our proposed corrective skip regularization on Human3.6m test set.
\ignore{All methods are initialized with the same weights, and trained on Human3.6m training data.} 
We observe that networks using vanilla skip connections perform markedly worse than the baseline, while corrective skip connections yield more than 5mm improvement for 7 classes of activities (marked as underlined).
\ignore{The same models are evaluated on~\dataset~test-set in Table \ref{tbl:our_test}. Additionally, corrective skip yields a significantly improved generalization to in-the-wild scenes, as evidenced by the $\approx$8\% 3DPCK improvement on outdoor sequences.}
\ignore{With unaugmented ~\dataset~training data, we again see $\approx8\%$ 3DPCK improvement on the outdoor sequences (Table \ref{tbl:our_test}) using corrective skip. The regularization interpretation is further supported by the decrease in performance on the studio sequences, which appear similar to training data. 
Augmented \dataset~training data has sufficient appearance variation for the the regularization effect of corrective skip to begin hurting performance, and the outdoor 3DPCK drops by $\approx12\%$ over the base network. 
Intuitively, corrective skip scheme changes the loss landscape for the core network, allowing it devote resources at training time to correcting larger mispredictions while the skip connections handle the vast numbers of mostly correct predictions which together present a steeper gradient on the loss landscape. This is similar to adaptive re-weighting in AdaBoost~\cite{adaboost}, emphasizing under-represented and difficult poses.}
We verified that the effect is not due to a higher effective learning rate seen by the core network due to the additional loss term. 
%Move to supplementary
\renewcommand{\tabcolsep}{1.5pt}
\begin{table}[]
\centering
\caption{Evaluation by scene-setting of our design choices on \dataset~test set with weight transfer from ImageNet. Training on our markerless dataset improves accuracy significantly, in particular with the proposed augmentation strategy. Fusion yields an additional gain. \emph{GS} indicates sequences with green screen. 
}
%All with perspective correction, and using ground truth bounding box crops
\label{tbl:our_test}
\resizebox{1\columnwidth}{!}{
\begin{tabular}{|c|lc|ccc|cc|}
\hline
  \multirow{3}{*}{3D dataset}  &   \multicolumn{2}{c|}{\multirow{3}{*}{Network architecture}}       & \multicolumn{1}{c|}{Studio}       & \multicolumn{1}{c|}{Studio}          &         \multirow{2}{*}{Outdoor} &                           \multicolumn{2}{c|}{\multirow{2}{*}{All}}      \\
&     \multicolumn{2}{c|}{}
&  \multicolumn{1}{c|}{GS} &  \multicolumn{1}{c|}{no GS} &  & \multicolumn{2}{c|}{}         \\ \cline{4-8}
&                   \multicolumn{2}{c|}{}        & \multicolumn{1}{c|}{\T 3DPCK}          & \multicolumn{1}{c|}{3DPCK}             & 3DPCK     & \multicolumn{1}{c|}{3DPCK}  & AUC  \\ \hline  \hline
\multicolumn{1}{|c|}{\multirow{2}{*}{Human3.6m}} & 
%\T Base          &  & 21.1         & 32.5            & 10.8    & \multicolumn{1}{c|}{22.6} & 8.8 \\ \cline{3-9}
%& & 
Base + Corr. Skip  \T &  & 22.2         & 33.9            & 18.5    & \multicolumn{1}{c|}{25.1} & 8.7 \\ \cline{2-8}
\multicolumn{1}{|c|}{} & \multicolumn{2}{l|}{Base + Corr. Skip + Fusion\T}     & 22.3         & 34.2            & 20.0    & \multicolumn{1}{c|}{26.0} & 9.5 \\ \hline 
\multicolumn{1}{|c|}{\multirow{2}{*}{Ours Unaug.}} & 
%\T Base          &  & 73.6         & 42.9            & 19.5    & \multicolumn{1}{c|}{49.0} & 23.3 \\ \cline{3-9}
%& & 
\T Base + Corr. Skip   &  & 66.9         & 38.2            & 27.9    & \multicolumn{1}{c|}{46.8} & 20.9 \\ \cline{2-8}
\multicolumn{1}{|c|}{} & \multicolumn{2}{l|}{Base + Corr. Skip + Fusion\T}   & 67.6         & 39.6            & 28.5    & \multicolumn{1}{c|}{47.8} & 21.8 \\ \hline 
\multicolumn{1}{|c|}{\multirow{2}{*}{Ours Aug.}}  & 
%\T Base     &  & 77.2         & 59.5            & 48.7    & \multicolumn{1}{c|}{63.7} & 31.1 \\ \cline{3-9}
\T Base + Corr. Skip     &  & 71.1         & 51.7            & 36.1    & \multicolumn{1}{c|}{55.4} & 26.0 \\ \cline{2-8}
  \multicolumn{1}{|c|}{}                 & \multicolumn{2}{l|}{Base + Corr. Skip + Fusion\T}       & \textbf{73.5}         & \textbf{53.1}            & \textbf{37.9}    & \multicolumn{1}{c|}{\textbf{57.3}} & \textbf{28.0} \\ \hline 
%&                   & Unug+Tx &  & 84.1         & 68.9            & 59.6    & \multicolumn{1}{c|}{72.5} & 36.9 \\
%&                   & Aug+Tx  &  & 82.6         & 66.7            & 62.0    & \multicolumn{1}{c|}{71.7} & 36.4
\end{tabular}
}
\vspace{-0.5cm}
\end{table}

%\paragraph{Multimodal prediction and fusion} 
\noindent\textbf{Multimodal prediction and fusion.}
The multi-modal fusion scheme yields noticeable improvement across all datasets tested in tables \ref{tbl:s9_11_w_imagenet} and \ref{tbl:our_test}. Since upright poses dominate in pose datasets, and the activity classes are often diluted significantly by upright poses, the true extent of improvement by the multi-modal fusion scheme is masked. To show that the fusion scheme indeed improves challenging pose classes, we cluster the Human3.6m test set by pose as shown in Figure \ref{fig:pose_clusters}, which visualizes the centroid of each cluster. 
Then we group the clusters visually into three pose classes, namely Stand/Walk, Sit and Crouch, going by the cluster representatives.
For the Stand/Walk class, adding fusion has minimal effect, going from 88.4mm to 88.8mm. However, for Sit class fusion leads to a 3.5mm improvement, from 118.9mm to 115.4mm. Similarly, Crouch class has the highest improvement of 5.5mm, going from 156mm to 150.5mm.
The improvement is not simply due to additional training, and is less pronounced if predicting $\Proot$, $\Ofirst$ and $\Osecond$ with a common stub, even with more features in the fully-connected layer. Details in the supplementary.
%Refer to the supplementary document for details.
%%!TEX root = ../article.tex
\begin{table}[]
\centering
\caption{Comparison of results on Human3.6m \cite{ionescu_human36_pami14} with the state of the art. Human3.6m, Subjects 1,5,6,7,8 used for training, and 9,11 used for testing. \textsuperscript{S} = Scaled to test subject specific skeleton, computed from T-pose. \textsuperscript{T}= Uses Temporal Information, \textsuperscript{J14/J17} = Joint set evaluated, \textsuperscript{A} = Uses Best Alignment To GT per frame, \textsuperscript{Act} = Activitywise Training, \textsuperscript{1/10/64} = Test Set Frame Sampling}
\label{tbl:s9_11_compare}
%\resizebox{1.8\columnwidth}{!}{
\resizebox{0.8\columnwidth}{!}{
\begin{tabular}{@{}lc@{}}
\multicolumn{1}{c|}{\textbf{Method}}                      & \textbf{Total MPJPE (mm)} \\ \hline \hline
\multicolumn{1}{l|}{Deep Kinematic Pose\cite{zhou_deep_kinematic_arxiv16}\textsuperscript{\textbf{J17},B}\T}      & 107.26  \\ \hline
\multicolumn{1}{l|}{Sparse. Deep. \cite{zhou_sparseness_deepness_cvpr15}\textsuperscript{T,\textbf{J17},B,10,Act}\T} & 113.01  \\\hline
\multicolumn{1}{l|}{Motion Comp. Seq. \cite{tekin_motion_comp_cvpr16}\textsuperscript{T,\textbf{J17},B}\T}       &  124.97  \\\hline
\multicolumn{1}{l|}{LinKDE \cite{ionescu_human36_pami14}\textsuperscript{\textbf{J17},B,Act}\T}                                    &           162.14 \\ \hline
\multicolumn{1}{l|}{Du et al. \cite{yu_mono_heightmap_eccv16}\textsuperscript{T,\textbf{J17},B}\T}                                    &           126.47 \\ \hline
\multicolumn{1}{l|}{Rogez et al. \cite{rogez_mocap_nips16}\textsuperscript{(\textbf{J13}),B,64}\T}       &  121.20  \\ \hline
\multicolumn{1}{l|}{SMPLify \cite{bogo_smpl_eccv16}\textsuperscript{\textbf{J14},B,A,(First cam.)}\T}       &  82.3 \\ \hline 
\multicolumn{1}{l|}{3D=2D+Matching \cite{chen_2d_match_cvpr17}\textsuperscript{\textbf{J17},B}\T}       &  114.18 \\ \hline 
\multicolumn{1}{l|}{Distance Matrix \cite{moreno_distance_matrix_cvpr17}\textsuperscript{\textbf{J17},B}\T}       &  87.30 \\ \hline 
\multicolumn{1}{l|}{Volumetric Coarse-Fine\cite{pavlakos_volumetric_cvpr17}\textsuperscript{\textbf{J17},B,S*}\T}       & 71.90  \\ \hline 
\multicolumn{1}{l|}{LCR-Net \cite{rogez_lcr_cvpr17}\textsuperscript{\textbf{J17},B}\T}       & 87.7  \\ \hline 
%\multicolumn{2}{l}{\textbf{Ours (with 2DPoseNet Transfer)}\T}                                                                                                                                                                              \\ \hline
%~~~Corr. Skip + Fusion
\multicolumn{1}{l|}{\T Full model (w/o \dataset) \textsuperscript{\textbf{J17},B}}                                                                &  74.11  \\ 
\multicolumn{1}{l|}{\T Full model (w/o \dataset) \textsuperscript{\textbf{J17},B,\textbf{S}}}                                                       & 68.61   \\ 
\multicolumn{1}{l|}{\T Full model (w/o \dataset) \textsuperscript{\textbf{J14},B,\textbf{A}}}                                                       & \textbf{54.59}   \\ \hline 
%\multicolumn{1}{l|}{Tekin et al \cite{tekin_structured_bmvc16}\textsuperscript{\textbf{J17},B,Act}}                              & -            & -       & 162.17       & -       & 65.75  & 130.53   & \multicolumn{1}{c|}{-}         & -       \\
%\multicolumn{1}{l|}{Li et al \cite{li_maxmargin_iccv15}}                                     & -            & -       & 168.68       & -       & 69.97   & 132.17   & \multicolumn{1}{r|}{-}         & -       \\
%\multicolumn{1}{l|}{Li et al \cite{li_accv14}\textsuperscript{\textbf{J17},B,Act}}                                                & -            & -       & 189.08       & -       & 77.60   & 146.59   & \multicolumn{1}{r|}{-}         & -       \\
\end{tabular}
}
\vspace{-0.5cm}
\end{table}

\mysubsection{Transfer Learning}{tx_learn}
\ignore{Despite the improved generalization brought about by our dataset and supervision techniques (Table \ref{tbl:our_test}) for 3D pose estimation, it doesn't approach the level of performance seen for 2D pose estimation methods. 
Our \emph{2DPoseNet} achieves 91.2 PCK and 66.3 AUC on the LSP test set, and 89.7 PCK and 61.3 AUC on MPII Single Person test set.
}
Our approach of transferring representations from \emph{2DPoseNet} to \emph{3DPoseNet} yields 64.7\% 3DPCK on~\dataset~test-set when trained with only Human3.6m data, compared to 63.7\% 3DPCK of the model trained on our augmented training set without transfer learning. It also shows state of the art performance on Human3.6m test set with an error of $\approx$74mm, demonstrating the dual advantage of the approach in improving both the accuracy of pose estimation and generalizability to in-the-wild scenes. Combining our dataset and transfer learning leads to the best results at $\approx72.5\%$ 3DPCK. See Table \ref{tbl:our_test_txlearn}.

In contrast to existing approaches countering data scarcity,
transfer learning does not require complex dataset synthesis,
yet exceeds the performance of Chen \etal~\cite{chen_synth_data_3dv16} (with synthetic data and domain adaptation, 28.8\% 3DPCK, after procrustes alignment) and our 
base model trained with the synthetic 
%CMU+MPII+LSP 
data of Rogez \etal~\cite{rogez_mocap_nips16} (21.7\% 3DPCK). 
Our approach also performs better than domain adaptation \cite{ganin_domain_icml15} to in-the-wild data (Table \ref{tbl:our_test_txlearn}). Details in the supplementary.
%.
\mysubsection{Benefit of \dataset}{dataset_results}
Evaluating on \dataset~test-set, without any transfer learning from \emph{2DPoseNet}, we see in Table \ref{tbl:our_test} that our dataset, even without augmentation, leads to a $\approx$9\% 3DPCK improvement on outdoor scenes over Human3.6m. 
However, our augmentation strategy is crucial for improved generalization, as seen from the gains in 3DPCK across scene settings in Table \ref{tbl:our_test}, giving $57.3\%$ 3DPCK overall.

Even when combined with transfer learning, we see in Table \ref{tbl:our_test_txlearn} that our dataset (both augmented and unagumented) consistently performs better than Human3.6m. The best performance of $76.5\%$ 3DPCK on \dataset~test set and of 72.88mm on Human3.6m is obtained when the two datasets are combined with transfer learning. 

\mysubsection{Other Components}{model-components}
\noindent\textbf{Bounding box computation.~}
On \dataset~test set, we additionally evaluate our best performing network using bounding boxes computed from \emph{2DPoseNet}. As shown in Table \ref{tbl:our_test_txlearn}, the performance drops to 74.4\%~3DPCK from 76.5\%~3DPCK due to the additional difficulty.

\noindent\textbf{Perspective correction.~}
Table \ref{tbl:our_test_txlearn} shows that perspective correction also has a significant impact, without which, the performance drops to 73\% 3DPCK from 76.5\%.
\renewcommand{\tabcolsep}{1.5pt}
\newcommand\Tstrut{\rule{0pt}{2.6ex}}       % "top" strut
\newcommand\Bstrut{\rule[-0.9ex]{0pt}{0pt}} % "bottom" strut
\newcommand{\TBstrut}{\Tstrut\Bstrut} % top&bottom struts
{
\vspace{0.1cm}
\begin{table}[]
\centering
\caption{Evaluation on \dataset~test set with weight transfer from \emph{2DPoseNet}, by scene setting.
Training our full model on our dataset paired with Human3.6m yields best accuracy over all. 
\emph{GS} indicates sequences with green screen background. 
%All have weights transferred from 2DPoseNet, have perspective correction applied, and use bounding box annotation, except where it is explicitly mentioned.
}
\label{tbl:our_test_txlearn}
\resizebox{1.0\columnwidth}{!}{
\begin{tabular}{|c|l|ccc|cc|}
\hline \multicolumn{1}{|c}{\multirow{3}{*}{3D dataset}}    & \multicolumn{1}{|c}{\multirow{3}{*}{Method}} & \multicolumn{1}{|c|}{Studio}      & \multicolumn{1}{|c|}{Studio}          &        \multirow{2}{*}{Outdoor}   &                 \multicolumn{2}{|c|}{\multirow{2}{*}{All}}               \Bstrut\\
                     \multicolumn{1}{|c}{}      & \multicolumn{1}{|c|}{} & \multicolumn{1}{c|}{GS} & \multicolumn{1}{c|}{no GS} & & \multicolumn{2}{|c|}{}          \\ \cline{3-7}
                     \multicolumn{1}{|c}{}       &   \multicolumn{1}{|c|}{}& \multicolumn{1}{c|}{\T 3DPCK}          & \multicolumn{1}{c|}{~3DPCK~}             & ~3DPCK~     & \multicolumn{1}{c|}{~3DPCK~}  & ~AUC~  \\ \hline  \hline
 \multirow{2}{*}{~Human3.6m}         &~Domain adapt.  & 44.1         & 42.6            & 35.2    & \multicolumn{1}{c|}{41.4} & 17.7 \TBstrut\\ \cline{2-7}
           & ~Ours (full model) & 70.8         & 62.3            & 58.5    & \multicolumn{1}{c|}{64.7} & 31.7 \TBstrut\\ \hline
% \T Ours.         & Base + Fusion & 82.8         & 68.0            & 62.3    & \multicolumn{1}{c|}{72.4} & 36.9 \\
  Ours Aug.         & ~Ours (full model) & 82.6         & 66.7            & 62.0    & \multicolumn{1}{c|}{71.7} & 36.4 \TBstrut\\ \hline
  ~Ours Unaug.         & ~Ours (full model) & 84.1         & 68.9            & 59.6    & \multicolumn{1}{c|}{72.5} & 36.9 \TBstrut\\ \hline
 Ours Aug.     & \multicolumn{1}{l|}{~Ours, w/o persp. corr.} & 81.9         & 68.6            & 67.4    & \multicolumn{1}{c|}{73.5} & 37.6 \TBstrut\\ 
\cline{2-7}
  +        & ~Ours, w/o GT BB & 80.4         & 71.2           & 69.8    & \multicolumn{1}{c|}{74.4} & 39.6 \TBstrut\\
 \cline{2-7}
 Human3.6m         &\multicolumn{1}{l|}{~Ours (full model)} & \textbf{84.6}         & \textbf{72.4}            & \textbf{69.7}    & \multicolumn{1}{c|}{\textbf{76.5}} & \textbf{40.8} \TBstrut\\  \hline
%&                   & Unug+Tx &  & 84.1         & 68.9            & 59.6    & \multicolumn{1}{c|}{72.5} & 36.9 \\
%&                   & Aug+Tx  &  & 82.6         & 66.7            & 62.0    & \multicolumn{1}{c|}{71.7} & 36.4
\end{tabular}
}
\vspace{-0.4cm}
\end{table}
}

\mysubsection{Quantitative Comparison}{3DResults}
\noindent\textbf{Human3.6m.} 
Table \ref{tbl:s9_11_compare} shows comparison of our method with existing methods, all trained on Human3.6m.
Altogether, with our supervision contributions and transfer learning, we are the state of the art (74.11mm, without scaling), while also generalizing to in-the-wild scenes. Note that the Volumetric coarse to fine approach \cite{pavlakos_volumetric_cvpr17} requires estimates of the bone lengths to convert their predictions from pixels to 3D space.  
Complementing {Human3.6m} with our augmented \dataset~dataset further reduces the error to 72mm. 

\noindent\textbf{HumanEva.} The improvements on Human3.6m are confirmed with a 30.8 and 33.5 MPJPE score on the S1 Box and Walk sequences of HumanEva, after alignment. See supplemental document.
\ignore{\paragraph{HumanEva}
In addition to evaluating centered pose $\Proot$,
we evaluate the global 3D pose prediction $\PG$ on the widely used HumanEva motion capture dataset --- \emph{Box} and \emph{Walk} sequences of \emph{Subject 1} from the validation set. Note that we do not use any data from HumanEva for training.
We significantly improve the state of the art for the Box sequence (82.1mm \cite{bogo_smpl_eccv16} vs 58.6mm). Results on the Walk sequence are of higher accuracy than Bogo \etal \cite{bogo_smpl_eccv16}, but lower than the accuracy of Bo \etal \cite{bo_twin_ijcv10} and Yasin \etal \cite{yasin_dual_source_cvpr16}, who, however train on HumanEva \cite{bo_twin_ijcv10} or use an example database dominated by walking motions \cite{yasin_dual_source_cvpr16}. Further, our skeletal structure does not match that of HumanEva, which contributes some of the error. 
See Table \ref{tbl::HumanEva}.
Our end to end run time is less than 1s.}

\noindent\textbf{\dataset.} 
\ignore{Our new \dataset~test set complements existing test sets with additional pose variation and appearance variation, in different settings. This makes it suitable for testing the generalization of various methods. We report PCK and AUC by activity classes in Table \ref{tbl:our_test_with_h36m}, alongside a similar grouping of activities on Human3.6m. We again see that 2DPoseNet transfer learning improves significantly on our baseline network, going from 22.5\%PCK to 60.9\%PCK. Improvements using the other proposed supervision schema are also evident. Complementing {Human3.6m} with our augmented dataset further improves performance.}
We also evaluated some of the existing methods on our test set. Deep Kinematic Pose 
\cite{zhou_deep_kinematic_arxiv16}, attains 13.8\% 3DPCK overall.
Our full model attains significantly higher accuracy: without transfer learning and trained on Human3.6m obtains 26\% 3DPCK, and 64.7\% 3DPCK with transfer learning.
%, and with combined data we are at 76.5\% 3DPCK.%Zhou et al \cite{zhou_deep_kinematic_arxiv16} perform X with wth a PCK of showing \TODO....
\ignore{incorporate this sentence somewhere: The error numbers as compared to Human3.6m are generally larger, this confirms the increased difficulty and generality of the proposed test set.}
The large discrepancy in performance between Human3.6m and our new in-the-wild test set highlights the importance of a new benchmark to test generalization to natural images and motions.

%
%!TEX root = ../article.tex
\mysection{Discussion}{sec:discussion}
Despite the demonstrated competitive results, our method and others have limitations. Most training sets, also \cite{chen_synth_data_3dv16}, have a strong bias towards chest height cameras. 
Thus, estimating 3D pose from starkly
different camera views is still a challenge. Our new dataset provides diverse view-points, which can support development towards viewpoint invariance in future methods. 
Similar to related
approaches, our per-frame estimation exhibits temporal jitter on video sequences. In future, we will investigate integration with model-based temporal tracking to further increase accuracy and
temporal smoothness. 
At less than 250\,ms per frame, our approach is much faster than model based methods which work offline in the order of minutes. There still remains scope for improvement towards real time, through smaller input resolution and shallower networks.
 
We also show that joining forces with transfer learning, in conjunction with algorithmic and data contributions, will aide progress in 3D pose estimation in many different directions, such as overall accuracy and generalizability. 
\mysection{Conclusion}{sec:discussion}
We have presented a fully feedforward CNN-based approach for monocular 3D human pose estimation that attains state-of-the-art on established benchmarks \cite{ionescu_human36_pami14, sigal_humaneva_ijcv10} and quantitatively outperforms existing methods on the introduced in-the-wild benchmark. 
State of the art is attained with enhanced CNN supervision techniques and improved parent relationships in the kinematic chain.
Transfer learning from in-the-wild 2D pose data in tandem with a new dataset that includes a larger variety of real and augmented human appearances, activities and camera views, leads to the significantly improved generalization to in-the-wild images.
Our method is also the first to efficiently extract global 3D position in non-cropped images, without time consuming iterative optimization.

%!TEX root = ../article.tex
%%%%%%%%% TITLE
%\title{Monocular 3D Human Pose Estimation\\ In The Wild Using Improved CNN Supervision}
\begin{center}
\textbf{\large Supplemental Document: Monocular 3D Human Pose Estimation\\ In The Wild Using Improved CNN Supervision}
\end{center}
\setcounter{equation}{0}
\setcounter{figure}{0}
\setcounter{table}{0}
\setcounter{section}{0}

\section*{This document accompanies the main paper, and the supplemental video.}

\section{Further Discussion of Design Choices Regarding Multi-modal Fusion}
To demonstrate that the improvement seen due to the fusion scheme is not simply a result of fine tuning, we compare the result of fusion with components successively removed. Using $\Proot$, $\Ofirst$ and $\Osecond$, we get an MPJPE of 74.49mm on Human3.6m. On removing $\Osecond$, the error increases to 74.77mm, and on removing both $\Ofirst$ and $\Osecond$, the error increases to 75.27mm. The comparison here is without any multi-level corrective skip training. 

For $\Proot$, $\Ofirst$ and $\Osecond$ to have different modes of mispredictions, the underlying feature set that they are computed from has to be as different as possible, because each is related to the other with a linear transform. 
We achieve some degree of decorrelation between the three by using 3 different prediction stubs, one each for $\Proot$, $\Ofirst$ and $\Osecond$ with a convolutional layer ($k_{5\times5}$, $s_2$) with 128 features followed by a fully-connected layer.
If we replace these three stubs with a single stub with the convolutional layer having 256 features followed by a fully-connected layer, the resulting MPJPE is 75.30mm after fusion, in contrast to an MPJPE of 74.49mm from fusing the result of 3 prediction stubs. Both of these are without corrective-skip connections.

\section{Further Discussion of Multi-level Corrective Skip}
Since our multi-level corrective skip scheme adds an additional loss at the last stage ($X_{\textrm{deep}}$, where $X$ is $P$/$O1$/$O2$) of the network, it increases the effective learning rate seen by the core network.
To verify that the improvements seen due to the proposed scheme are not caused by this difference in the effective learning rate, we trained a version of the \emph{Base} network with loss weights as the sum of the loss weights for $X_{\textrm{deep}}$ and $X_{\textrm{sum}}$ specified in Table \ref{tbl:3dposenet_taper_scheme}. 
We find that this network performs worse than the \emph{Base} network (107.14mm vs 104.32mm MPJPE on Human3.6m), and does not approach the accuracy attained with multi-level corrective skip scheme (101.09mm).

\section{Global Pose Computation}
\subsection{3D localization}

In this section we describe a simple, yet very efficient, method to compute the global 3D location $T$ of a noisy 3D point set $\Proot$ with unknown global position. We assume known scaling and orientation parameters, obtained from its 2D projection estimate $\Pprojected$ in a camera with known intrinsics parameters (focal length $f$).
We further assume that the point cloud spread in depth direction is negligible compared to its distance $z_0$ to the camera and
approximate perspective projection of an object near position $(x_0, y_0, z_0)^\top$ with weak perspective projection (linearizing the pinhole projection model at $z_0$):
\begin{equation}
\begin{pmatrix}u\\v\end{pmatrix} 
= \mathbf{\Pi} \begin{pmatrix}x\\y\\z\end{pmatrix}
\text{, with }
\mathbf{\Pi} =
\left(\begin{matrix}
\frac{f}{z_0}&0&0\\0&\frac{f}{z_0}&0 
\end{matrix}\right).
\end{equation}
Estimates $\Pprojected$ and $\Proot$ are assumed to be noisy due to estimation errors. 
We find the optimal global position $T$ in the least squares sense, by minimizing $T = \arg\min_{(x, y, z)} E(x, y, z)$, with
\begin{align}
E &= \sum_i \lVert\Pprojected^i - \Pi\left( (x, y, z)^\top + \Proot^i\right)\rVert^2 \nonumber\\
   &= \sum_i \lVert\Pprojected^i - \frac{f}{z} \left( (x, y)^\top + \Proot^i_{[xy]}\right)\rVert^2,
\end{align}
where $P^i$ and $K^i$ denote the $i$th joint position in 3D and 2D, respectively,
and $\Proot^i_{[xy]}$ the $xy$ component of $\Proot^i$.
It has partial derivative
\begin{equation}
\frac{\partial E}{\partial x} = \frac{2 f}{z} \sum_i K^i_{[x]} + \frac{f}{z} \left( P^i_{[x]} - x\right),
\end{equation}
where $\Proot_{[x]}$ denotes the $x$ part of $\Proot$, and $\bar{\Proot}$ the mean of $\Proot$ over all joints.
Solving $\frac{\partial E}{\partial x}=0$ gives the unique closed-form solutions 
$x = \bar{\Pprojected}_{[x]} \frac{z}{f}  - \bar{\Proot}_{[x]}$ and equivalently $y = \bar{\Pprojected}_{[y]} \frac{z}{f}  - \bar{\Proot}_{[y]}$,
for $\frac{\partial E}{\partial y}=0$.

Substitution of $x$ and $y$ in $E$ and differentiating with respect to $z$ yields
\begin{align}
\frac{\partial E}{\partial z} =&
\frac{f \sum_i (\Pprojected^i - \bar{\Pprojected})^\top (\Proot^i_{[xy]} - \bar{\Proot}_{[xy]})}{z^2}  \nonumber\\
&+ \frac{f^2 \sum_i \lVert \Proot^i_{[xy]} - \bar{\Proot}_{[xy]} \rVert^2}{z^3} .
\end{align}
Finally, solving $\frac{\partial E}{\partial z}=0$ gives the depth estimate
\begin{align}
z &= f \frac{\sum_i \lVert\Proot^i_{[xy]} - \bar{\Proot}_{[xy]}\rVert^2 }{\sum_i (\Pprojected^i - \bar{\Pprojected})^\top (\Proot^i_{[xy]} - \bar{\Proot}_{[xy]})}\nonumber\\
&\approx f \frac{\sqrt{\sum_i \lVert\Proot^i_{[xy]} - \bar{\Proot}_{[xy]}\rVert^2} }{\sqrt{\sum_i \lVert\Pprojected^i - \bar{\Pprojected} \rVert^2 }}
\text{,}
\end{align}
where $(\Pprojected^i - \bar{\Pprojected}) (\Proot^i - \bar{\Proot}) = \lVert \Pprojected^i - \bar{\Pprojected}\rVert \lVert \Proot^i - \bar{\Proot} \rVert \cos(\theta)$ is approximated for $\theta \approx 0$. 
This is a valid assumption in our case, since the rotation of 3D and 2D pose is assumed to be matching.

\paragraph{Evaluation on HumanEva:~}
In addition to evaluating centered pose $\Proot$,
we evaluate the global 3D pose prediction $\PG$ on the widely used HumanEva motion capture dataset --- \emph{Box} and \emph{Walk} sequences of \emph{Subject 1} from the validation set. Note that we do not use any data from HumanEva for training.
%To overcome the scale ambiguity, only the camera focal length as well as the subject height needs to be calibrated.
We significantly improve the state of the art for the Box sequence (82.1mm \cite{bogo_smpl_eccv16} vs 58.6mm). Results on the Walk sequence are of higher accuracy than Bogo \etal \cite{bogo_smpl_eccv16}, but lower than the accuracy of Bo \etal \cite{bo_twin_ijcv10} and Yasin \etal \cite{yasin_dual_source_cvpr16}, who, however train on HumanEva \cite{bo_twin_ijcv10} or use an example database dominated by walking motions \cite{yasin_dual_source_cvpr16}. 
Our skeletal structure does not match that of HumanEva, e.g. the head prediction has a consistent frontal offset and the hip is too wide.
To compensate, we compute a linear map of dimension 14x14 (number of joints) that maps our joint positions as a linear combination to the HumanEva structure.
The same mapping is applied at every frame, but is computed only once, jointly on the Box and Walk sequence, to limit the correction to global inconsistencies of the skeleton structure.
This fine-tuned result is marked by $\sim$ in Table~\ref{tbl::HumanEva}.
%

%!TEX root = ../../Supplemental/supplemental.tex
\begin{table*}[t]
	\centering
	\caption{Quantitative evaluation on HumanEva-I~\cite{sigal_humaneva_ijcv10}, with different alignment strategies used in the literature. 
		For reference, we also show multi-view existing results. Our models use no data from HumanEva for training, while the other methods listed train/finetune on HumanEva-I. * = Does not use GT Bounding Box information. $\dagger$ = Translation alignment only. $\sim$ = trained or fine-tuned on HumanEva-I.}
	\label{tbl::HumanEva}
	\resizebox{1.8\columnwidth}{!}{
		\newcommand*\rot{\rotatebox{90}}
		\begin{tabular}{|c|l|c|c|c|c|c|c|}
			\hline
			\cline{2-8}
			&	& \multicolumn{3}{c|}{S1 Box}    & \multicolumn{3}{c|}{S1 Walk} \\ \hline
			&	& \begin{tabular}{@{}c@{}}$\PG$\\(global)\end{tabular} & \begin{tabular}{@{}c@{}}$\Proot$\\(align\textsuperscript{S,T})\end{tabular}  & \begin{tabular}{@{}c@{}}$\Proot$\\(align\textsuperscript{R,S,T})\end{tabular} & \begin{tabular}{@{}c@{}}$\PG$\\(global)\end{tabular} & \begin{tabular}{@{}c@{}}$\Proot$\\(align\textsuperscript{S,T})\end{tabular}  & \begin{tabular}{@{}c@{}}$\Proot$\\(align\textsuperscript{R,S,T})\end{tabular}   \\ \cline{1-8} 
			\multirow{8}{*}{\scriptsize\rot{Monocular}} 
%			&	Our full model*           & \bf{129.5}  & \bf{69.4}       & \bf{58.6}    & \bf{145.6}   & \bf{79.2}  & \bf{67.1}     \\
%			&	w/o Persp. correct.* & 133.9  & 79.4       & 58.6    & 147.7   & 83.6   & 67.2     \\
			&	Our full model*           & 117.1 &   80.5  & 58.6    & 121.1 &  81.0  & 67.2     \\
			&	w/o Persp. correct.* &  116.1  & 79.4 &  58.6     & 123.9 &  83.6 &  67.3     \\
			&	Our full model*$\sim$ & \bf{77.9} &  \bf{38.4} & \bf{30.8} & \bf{89.1}  & \bf{48.2} &  \bf{33.5}    \\\cline{2-8} 
			&	Zhou  \etal  \cite{zhou_sparseness_deepness_cvpr15}$\sim$   & -      & -       & -       & -       & -   & 34.2        \\  % CNN is finetuned
			&	Bo \etal \cite{bo_twin_ijcv10}*$\sim$              & -      & -           & -    & -       & 54.8$\dagger$      & -     \\
			&	Yasin \etal \cite{yasin_dual_source_cvpr16}*$\sim$              & -      & -           & -    & 52.2       & -       & -     \\ % trained on HumanEva
			&	Bogo \etal \cite{bogo_smpl_eccv16}$\sim$              & -      & -           & 82.1    & -       & -       & 73.3     \\ % regression is fine tuned
			&	Akhter \etal \cite{akhter_pose_conditioned_cvpr15}  & -      & -           & 165.5   & -       & -       & 186.1    \\ % unknown, from Bogo
			&	Ramakris.  \etal  \cite{ramakrishna2012reconstructing}     & -      & -           & 151.0   & -       & -       & 161.8    \\ \cline{1-8}% unknown, from Bogo
%			&	Zhou  \etal  \cite{zhou_convexrelaxation_cvpr2015}$\sim$   & -      & 112.5       & -       & -       & 100.0   & -        \\ \cline{1-8} % can't find the result, from bogo et al.
			% 3D stuff
			\multirow{3}{*}{\scriptsize\rot{Mulit-view\ \ }} &	Amin \etal \cite{amin_multi_bmvc2013}    & \bf{47.7}   & -           & -       & \bf{54.5}    & -       & -        \\
			&	Rhodin  \etal   \cite{rhodin_general_eccv16}        & 59.7   & -           & -       & 74.9    & -       & -        \\
			&	Elhayek  \etal   \cite{elhayek_efficient_cvpr2015}        & 60.0   & -           & -       & 66.5    & -       & -        \\\cline{1-8} 
		\end{tabular}
	}
\end{table*}

\begin{figure}
	\centering
	\includegraphics[width=0.9\linewidth]{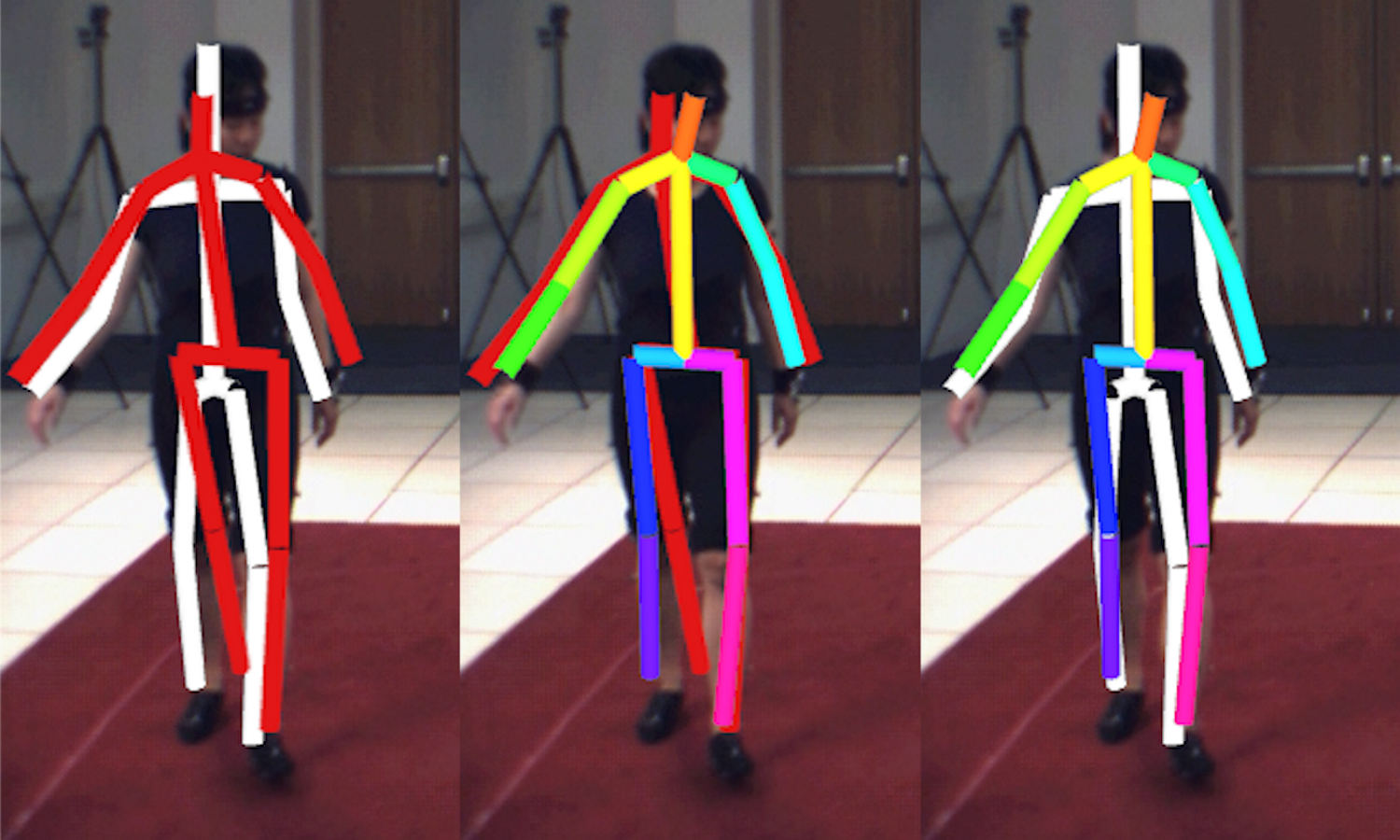}
	\caption{The predicted pose (red) is inaccurate for positions away from the camera center (left), compared against the ground truth (white).
		Perspective correction (colored) corrects the orientation (center) and is closer to the ground truth (right).
		Here tested on the walking sequence of HumanEva S1.}
	\label{fig:perspective_correction}
\end{figure}

\begin{figure}
	\centering
	\includegraphics[width=0.9\linewidth,trim={0 13.cm 20.cm 0},clip]{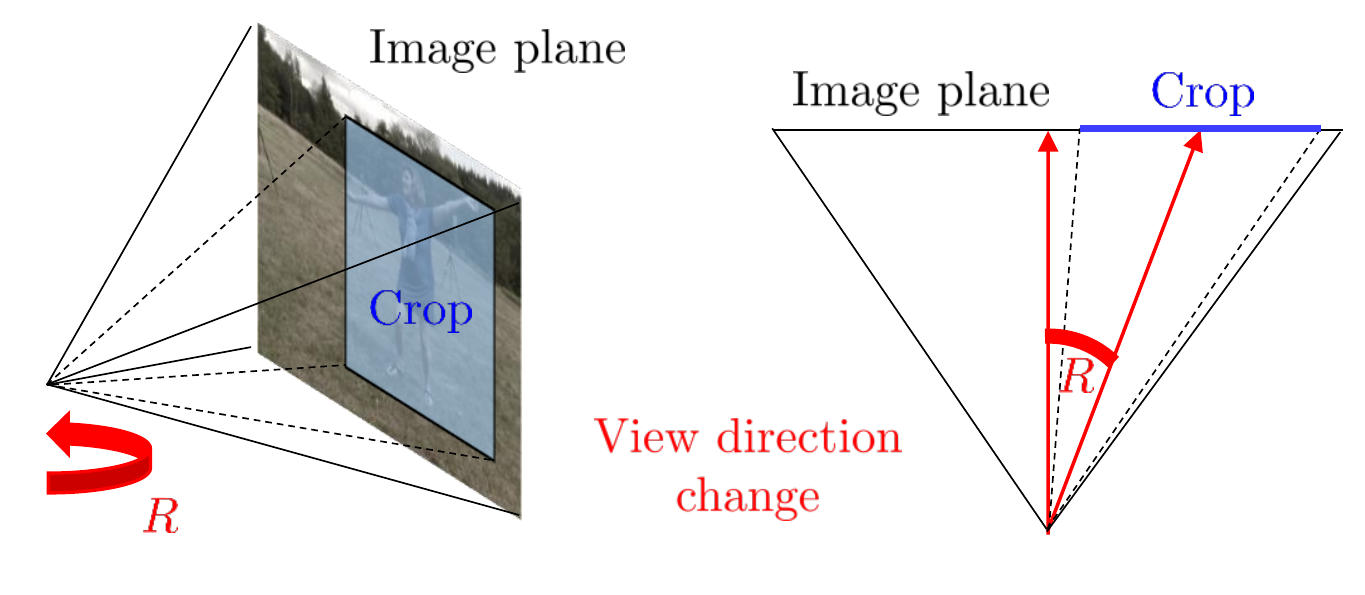}
	\caption{Sketch of the input image cropping and resulting change of field of view. The corresponding rotation $R$ of the view direction is sketched in 2D on the right.}
	\label{fig:cameraRotation}
\end{figure}

\subsection{Perspective correction}

Our \textit{3DPoseNet} predicts pose $\Proot$ in the coordinate system of the bounding box crop, which leads to inaccuracies as shown in Figure \ref{fig:perspective_correction}.
The cropped image appears if as it was taken from a virtual camera with the same origin as the original camera, but with view direction to the crop center, see Figure \ref{fig:cameraRotation}.
To map the reconstruction from the virtual camera coordinates to the original camera, we rotate $\Proot$ by the rotation
$R$ between the virtual and original camera. 
Since the existing training sets provide chest-height camera placements with the same viewpoint, the bias in vertical direction is already learned by the network. 
We apply perspective correction only in horizontal direction, where a change in cropping and yaw rotation of the person cannot be distinguished by the network.
R is then the rotation around the camera up direction by the angle between the original and the virtual view direction, see Figure \ref{fig:cameraRotation}.
On our \dataset~test set perspective correction improves the PCK by 3 percent points. 
On HumanEva the improvement is up to 3 mm MPJPE, see Table \ref{tbl::HumanEva}.
The correction is most pronounced for cameras with a large field of view, \eg GoPro and similar outdoor cameras, and when the subject is located at the border of the view.
Using the vector from the camera origin to the centroid of 2D keypoints $\Pprojected$ as the virtual view direction was most accurate in our experiments.
However, the crop center can be used instead. Opposed to the Perspective-n-Point algorithm applied by Zhou \etal \cite{zhou_sparseness_deepness_cvpr15}, 
any regression method that works on cropped images could immediately profit from this perspective correction, without computing 2D keypoint detections.

\begin{table}[]
	\centering
	\caption{Results of our \textit{2DPoseNet} on MPII Single Person Pose \cite{andriluka_mpii2d_cvpr14} dataset and LSP \cite{johnson_lsp_bmvc10} 2D Pose datasets. * = Trained/Finetuned only on the corresponding training set}
	\label{tbl:2dxpose}
	\begin{tabular}{ll||cc||cc} 
		%Only Results from 2016. Maybe put all in the Supplementary.
		\multicolumn{2}{l||}{}  & \multicolumn{2}{c||}{MPII} & \multicolumn{2}{c}{LSP}\\
		\multicolumn{2}{l||}{} & PCK$_{h0.5}$ & AUC & PCK$_{0.2}$ & AUC\\ \hline \hline
		\multicolumn{2}{l||}{Our \textit{2DPoseNet}} &  &  & &  \\
		%\multicolumn{2}{r||}{Base} & 89.5 & 61.3 &90.7 & 65.2 \\
		%\multicolumn{2}{r||}{+ MLC Sum} & 89.5 & \underline{61.6}  & 90.9 & \underline{65.7} \\  
		\multicolumn{2}{r||}{w Person Locali.} & \underline{89.7} & 61.3  & \textbf{91.2} & 65.3 \\
		\multicolumn{2}{r||}{w/o Person Locali.} & 89.6 & 61.5  & \textbf{91.2} & 65.5 \\ \hline 
		\multicolumn{2}{l||}{Stacked Hourgl.\cite{newell_stacked_hourglass_eccv16}} & \textbf{90.9}* & \textbf{62.9}*  & -&- \\
		\multicolumn{2}{l||}{Bulat et al.\cite{bulat_convpart_eccv16}}& 89.7* & 59.6*  & \underline{90.7} & - \\
		\multicolumn{2}{l||}{Wei et al.\cite{wei_cpm_cvpr16} }& 88.5 & 61.4  &90.5 &65.4 \\
		\multicolumn{2}{l||}{DeeperCut \cite{insafutdinov_deepercut_eccv16}}& 88.5 & 60.8 & 90.1&\textbf{66.1}  \\
		\multicolumn{2}{l||}{Gkioxary et al \cite{gkioxari_chained_eccv2016}}& 86.1* & 57.3* & -&-  \\
		\multicolumn{2}{l||}{Lifshitz et al. \cite{lifshitz_deep_consensus_eccv16} }& 85.0 & 56.8  & 84.2 & - \\
		\multicolumn{2}{l||}{Belagiannis et al.\cite{belagiannis_recurrent_arxiv6}}& 83.9* & 55.5*  & 85.1 & -\\
		\multicolumn{2}{l||}{DeepCut\cite{pishchulin_deepcut_cvpr16}}& 82.4 & 56.5 & 87.1& 63.5  \\
		\multicolumn{2}{l||}{Hu\&Ramanan \cite{hu_bottomup_cvpr16}}& 82.4* & 51.1* & - & -  \\
		\multicolumn{2}{l||}{Carreira et al. \cite{carreira_iterative_cvpr16}}& 81.3* & 49.1*  & 72.5* &- \\
	\end{tabular}
\end{table}
\section{CNN Architecture and Training Specifics}
\subsection{2DPoseNet}
\noindent\textbf{Architecture:~}
The architecture derives from Resnet-101, using the same structure as is until level 4. Since we are interested in predicting heatmaps, we remove striding at level 5. Additionally, the number of features in the \texttt{res5a} module are halved, identity skip connections are removed from \texttt{res5b} and \emph{res5c}, and the number of features gradually tapered to 15 (heatmaps for 14 joints + root). 
As shown in Table \ref{tbl:2dxpose}, for \textit{2DPoseNet}, our results on MPII and LSP test sets approach that of the state of the art. 

\noindent\textbf{Intermediate Supervision:~} Additionally, we employ intermediate supervision at \texttt{res4b20} and \texttt{res5a}, treating the first 15 feature maps of the layers as the intermediate joint-location heatmaps. Further, we use a Multi-level Corrective Skip scheme, with skip connections coming from \texttt{res3b3} and \texttt{res4b22} through prediction stubs comprised of a $1\times1$ convolution with 20 feature maps followed by a $3\times3$ convolution with 15 outputs.

\noindent\textbf{Training:~}For training, we use the Caffe~\cite{jia_caffe_ICM} framework, with the AdaDelta solver with a momentum of 0.9 and weight decay rate of 0.005.  We employ a batch size of 7, and use Euclidean Loss everywhere. For the Learning Rate and Loss Weight taper schema, refer to Table \ref{tbl:2dposenet_taper_scheme}.

%!TEX root = ../supplemental.tex
\begin{table}[]
\centering
\caption{Loss weight and learning rate, LR, taper scheme used for \textit{2DPoseNet}. \textit{2DPoseNet} also employs Multi-level Corrective Skip connections, and the heatmap $H_{\text{sum}}$ is the sum of $H_{\text{deep}}$ and the skip connections. Heatmaps $H_{\text{4b20}}$ and $H_{\text{5a}}$ are used for intermediate supervision.}
\label{tbl:2dposenet_taper_scheme}
\begin{tabular}{|c|c|c|c|c|c|}
\hline
Base & \multirow{2}{23px}{\# Iter}& \multicolumn{4}{c|}{Loss Weights ($\mathbf{w} \times L(H_{xx})$)} \\
LR  &  & $H_{\text{sum}}$ & $H_{\text{deep}}$ & $H_{\text{4b20}}$ & $H_{\text{5a}}$  \\ \hline
0.050     & 60k    & 1.0 & 0.5     & 0.5     & 0.5    \\
0.010    & 60k    & 1.0 & 0.4     & 0.1     & 0.1    \\
0.005    & 60k    & 1.0 & 0.2     & 0.05    & 0.05   \\
0.001    & 60k    & 1.0 & 0.2     & 0.05    & 0.05   \\
6.6e-4  & 60k    & 1.0 & 0.1     & 0.005   & 0.005  \\
0.0001   & 40k    & 1.0 & 0.01    & 0.001   & 0.001  \\
2.5e-5 & 40k    & 1.0 & 0.001   & 0.0001  & 0.0001 \\ \hdashline
0.0008   & 60k    & 1.0 & 0.0001  & 0.0001  & 0.0001 \\
0.0001   & 40k    & 1.0 & 0.0001  & 0.0001  & 0.0001 \\
3.3e-5 & 20k    & 1.0 & 0.0001  & 0.0001  & 0.0001 \\ \hline
\end{tabular}
\end{table}

\subsection{3DPoseNet}
\noindent\textbf{Architecture:~}
The core network is identical to \emph{2DPoseNet} up to \texttt{res5a}. A 3D Prediction stub is attached on top, comprised of a $5\times5$ convolution layer with a stride of 2 and 128 features, followed by a fully-connected layer.

\noindent\textbf{Multi-level Corrective Skip:~} 
We attach 3D prediction stubs to \texttt{res3b3} and \texttt{res4b20}, similar to the final prediction stub, but with 96 convolutional features instead of 128. The resulting predictions are added to $\PD$ to get $\PS$. We add a loss term to $\PD$ in addition to the loss term at $\PS$.

\noindent\textbf{Multi-modal Fusion:~} 
We add prediction stubs for $\Ofirst$ and $\Osecond$, similar to those for $\Proot$. Note that the predictions for $\Proot$, $\Ofirst$ and $\Osecond$ are done with distinct stubs, and this slight decorrelation of predictions is important. These predictions are at a later finetuning step fed into three fully-connected layers, with 2k, 1k and 51 nodes respectively.

\begin{table}[]
	\centering
	\caption{Loss weight and LR taper scheme used for \textit{3DPoseNet}. There is a difference in the number of iterations used when training with Human3.6m or \dataset~alone, v.s. when training with the two in conjunction. Part Labels $PL$ are used only when training with H3.6m solely. Multi-level skip connections add up with $X_{deep}$ to yield $X_{sum}$, where $X$ is $\Proot$ or $\Ofirst$ $\Osecond$.}
	\label{tbl:3dposenet_taper_scheme}
	\resizebox{1\columnwidth}{!}{
		\begin{tabular}{|l|c|c|c|c|c|c|c|c|}
			\multicolumn{9}{c}{}\\
			\hline
			& H3.6m/Our & H3.6m+Our& \multicolumn{6}{c|}{Loss Weights (w $\times$ L($A_{bb}$))}                                              \\ 
			Base  & Batch = 5 & Batch = 6 & \multicolumn{4}{c|}{$X = P/O1/O2$}    & \multicolumn{2}{c|}{}                                          \\ \cdashline{2-3}
			LR    & \#Epochs    & \#Epochs     & $X_{4b5}$ & $X_{4b20}$   & X$_\text{deep}$ & \multicolumn{1}{l|}{$X_{sum}$}& $H$     & $PL$*   \\ \hline
			0.05  & 3 (45k)               & 2.4 (60k)             & 50     & 50        & 50            & \multicolumn{1}{l|}{100} & 0.1   & 0.05   \\
			0.01  & 1 (15k)               & 1.2 (30k)             & 10     & 10       & 10            & \multicolumn{1}{l|}{100} & 0.05  & 0.025   \\
			0.005 & 2 (30k)               & 1.2 (30k)             & 5      & 5        & 5             & \multicolumn{1}{l|}{100}  & 0.01  & 0.005  \\
			0.001 & 1 (15k)               & 0.6 (15k)             & 1      & 1        & 1             & \multicolumn{1}{l|}{100}  & 0.01  & 0.005  \\
			5e-4  & 2 (30k)               & 1.2 (30k)             & 0.5    & 0.5      & 0.5           & \multicolumn{1}{l|}{100} & 0.005 & 0.001   \\
			1e-4  & 1 (15k)               & 0.6 (15k)             & 0.1    & 0.1      & 0.1           & \multicolumn{1}{l|}{100} & 0.005 & 0.001   \\ \hline
		\end{tabular}
	}
\end{table}

\noindent\textbf{Intermediate Supervision:~} 
We use intermediate supervision at \texttt{4b5} and \texttt{res4b20}, using prediction stubs comprised of $7\times7$ convolution with a stride of 3 and 128 features, followed by a fully-connected layer predicting $\Proot$, $\Ofirst$ and $\Osecond$ as a single vector. Additionally, we predict joint location heatmaps and part-label maps using a $1\times1$ convolution layer after \texttt{res5a} as an auxiliary task. We don't use the part-label maps when training with \dataset~dataset.

\noindent\textbf{Training:~}
For training, the solver settings are similar to \emph{2DPoseNet}, and we use Euclidean Loss everywhere. For transfer learning, we scale down the learning rate of the transferred layers by a factor determined by validation. For fine-tuning in the multi-modal fusion case, we similarly downscale the learning rate of the trained network by 10,000 with respect to the three new fully-connected layers. For the learning rate and loss weight taper schema for both the main training and multi-modal fusion fine-tuning stages, refer to Tables \ref{tbl:3dposenet_taper_scheme} and \ref{tbl:3dposenet_finetune_scheme}. We use different training durations when using Human3.6m or \dataset~in isolation, versus when using both in conjunction. This is reflected in the aforementioned tables.

\begin{table}[t!]
	\centering
	\caption{Loss weight and LR taper scheme used for fine tuning \textit{3DPoseNet} for Multi-modal Fusion scheme.  }
	\label{tbl:3dposenet_finetune_scheme}
	\resizebox{1\columnwidth}{!}{
	\begin{tabular}{|l|c|c|c|}
		\multicolumn{4}{c}{}\\
		\hline
		& H3.6m/Our & H3.6m+Our& \multicolumn{1}{c|}{}                                              \\ 
		Base  & Batch = 5 & Batch = 6 & \multicolumn{1}{c|}{Loss Weights (w $\times$ L($A_{bb}$))}   \\ \cdashline{2-3}
		LR    & \#Epochs    & \#Epochs     & $\PF$    \\ \hline
		0.05  & (1k)               &  (2k)             & 100     \\
		0.01  & 1 (15k)               & 0.8 (20k)             & 100     \\
		0.005 & 1 (15k)               & 0.8 (20k)             & 100       \\
		0.001 & 1 (15k)               & 0.8 (20k)             & 100       \\ \hline
	\end{tabular}}
    \vspace{-0.5cm}
\end{table}

\subsubsection{3D Pose Training Data}
In the various experiments on \emph{3DPoseNet}, for the datasets we consider, we select $\approx$37.5k frames for each, yielding $\approx$75k samples after scale augmentation at 2 scales (0.7 and 1.0).

\noindent\textbf{Human3.6m:}
We use the H80k~\cite{ionescu_iterated_cvpr14} subset of Human3.6m, and train with the ``universal" skeleton, using subjects S1,5,6,7,8 for training and S9,11 for testing. 
The predicted skeleton is not scaled to the test subject skeletons at test time.
%\paragraph{\dataset}

\begin{figure*}
	\centering
	\includegraphics[width=1.980\columnwidth]{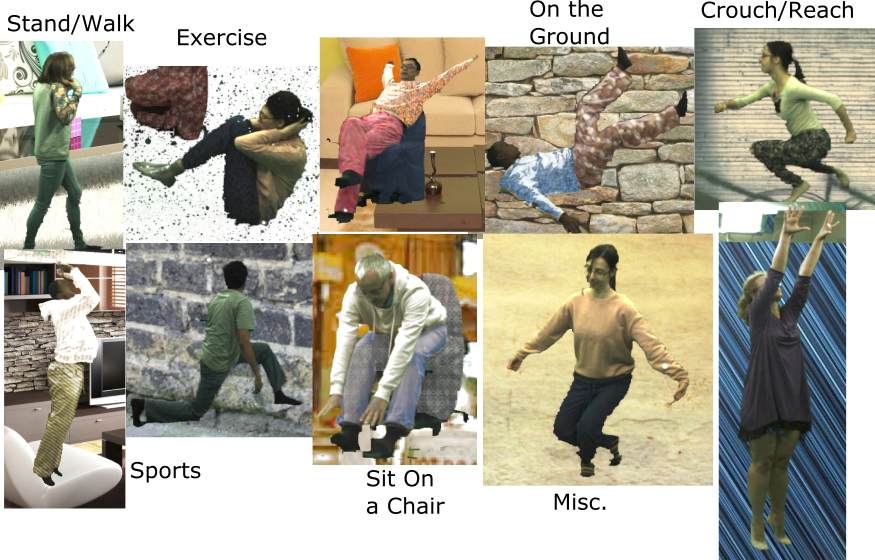}
	\caption{A sample of the activities, clothing, subjects as well as augmentation on ~\dataset~Trainig Set.}
	\label{fig:our_dataset_train_set_activites}
\end{figure*}
\begin{figure*}
	\centering
	\includegraphics[width=1.980\columnwidth]{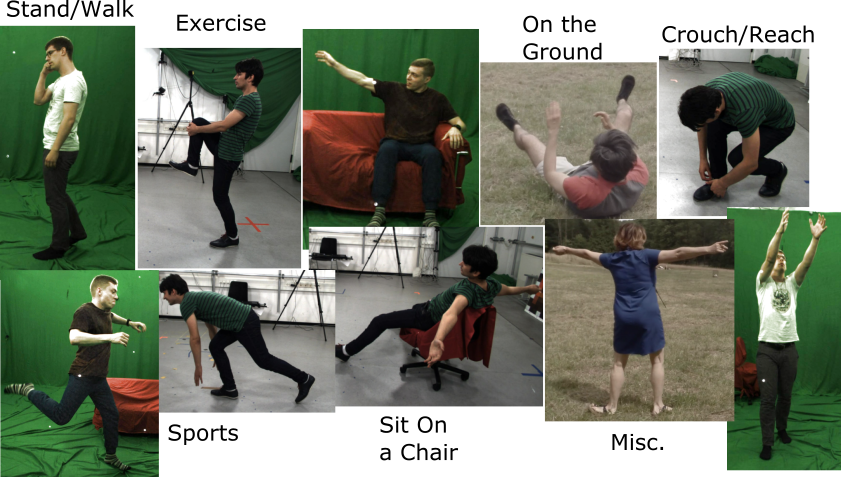}
	\caption{A sample of the activities and subjects in the test set of \dataset}
	\label{fig:our_dataset_test_set_activites}
\end{figure*}
\noindent\textbf{\dataset:}
For our dataset, to maintain compatibility of view with Human3.6m and other datasets, we only pick the 5 chest high cameras for all 8 subjects, sampling frames such that at least one joint has moved by more than 200mm between selected frames. A random subset of these frames is used for training, to match the number of selected Human3.6m frames.
\ignore{, giving us 500k frames. We sample frames such that at least one joint has moved by more than 200mm between selected frames. From the resulting 100k frames, we randomly sample 37.5k frames, and get 75k frames after scale augmentation.}
%\paragraph{\dataset~Augmented}

\noindent\textbf{\dataset~Augmented:}
The augmented version uses the same frames as the unaugmented~\dataset~above, keeping $\approx$25\% frames unaugmented, $\approx$40\% with only BG and Chair augmentation, and the rest with full augmentation.

\subsubsection{Domain Adaptation To In The Wild 2D Pose Data}
We use a domain adaptation stub comprised of $conv_{3\times3,256}$, $conv_{3\times3,128}$, $fc_{64}$ and $fc_{1}$ layers, and cross entropy domain classification loss. It uses Ganin \etal's \cite{ganin_domain_icml15} gradient inversion approach. The domain adaptation stub is attached after $res4b22$ in the network. We found that directly starting out with $\lambda=-1$ performs better than gradually increasing the magnitude of $\lambda$ with increasing iterations. We train on the Human3.6m training set, with 2D heatmap and part label prediction as auxiliary tasks. Images from MPII \cite{andriluka_mpii2d_cvpr14} and LSP \cite{johnson_lsp_bmvc10,johnson_lspet_cvpr11} training sets are used without annotations for learning better generalizable features. The generalizability is improved, as evidenced by the 41.4 3DPCK on \dataset~test set, but does not match up with the 64.7 3DPCK attained using transfer learning. Detailed results in main Table 3.

\section{\dataset~Dataset}
We cover a wide range of poses in our training and test sets, roughly grouped into various activity classes. A detailed description of the dataset is available in Section 4 of the main paper. In addition, Figure \ref{fig:our_dataset_train_set_activites} samples the various different activity classes, augmentation and subjects represented in our dataset.

Similarly for the test set, we show a sample of the activities and the variety of subjects in Figure \ref{fig:our_dataset_test_set_activites}.

\subsection{The Challenge of Learning Invariance to Viewpoint Elevation}
In this paper, we only consider the cameras in the training set placed at chest-height, in part to be compatible with the existing datasets, and in part because viewpoint elevation invariance is a significantly more challenging problem. Existing benchmarks do not place emphasis on this. We will release an expanded version of our \dataset~ testset with multiple camera viewpoint elevations, to complement the training data.

{\small
\bibliographystyle{ieee}
\bibliography{article}
}

{\small
\bibliographystyle{ieee}
\bibliography{article}
}

\end{document}